\documentclass{article}

\usepackage{aistats2018}
\usepackage[utf8]{inputenc} 
\usepackage[T1]{fontenc}    
\usepackage{hyperref}       
\usepackage{url}            
\usepackage{booktabs}       
\usepackage{amsfonts}       
\usepackage{nicefrac}       
\usepackage{microtype}      
\usepackage{balance}
\usepackage{graphicx} 
\usepackage{amsmath,amssymb,amsfonts,amstext,amsthm,bm}
\usepackage{latexsym,graphics,epsf,epsfig,psfrag}
\usepackage{color}
\usepackage{thmtools}
\usepackage{algorithm}
\usepackage{algorithmicx}
\usepackage[noend]{algpseudocode}
\usepackage{dsfont}
\usepackage{enumitem}
\usepackage{cleveref}

\usepackage{natbib}
\usepackage{framed}

\newtheorem{proposition}{Proposition}

\newtheorem{thm}{Theorem}
\newtheorem{coro}{Corollary}

\newcommand{\Ab}{\bm{A}}
\newcommand{\Bb}{\bm{B}}

\newcommand{\Cb}{\bm{C}}
\newcommand{\Qb}{\bm{Q}}
\newcommand{\Gb}{\bm{G}}
\newcommand{\Ib}{\bm{I}}

\newcommand{\Hb}{\bm{H}}
\newcommand{\Ub}{\bm{U}}
\newcommand{\Vb}{\bm{V}}
\newcommand{\Wb}{\bm{W}}
\newcommand{\Xb}{\bm{X}}
\newcommand{\Sigmab}{\bm{\Sigma}}
\newcommand{\Yb}{\bm{Y}}

\newcommand{\eb}{\mathbf{e}}
\newcommand{\xb}{\mathbf{x}}
\newcommand{\ab}{\mathbf{a}}

\newcommand{\wb}{\mathbf{w}}
\newcommand{\ub}{\mathbf{u}}
\newcommand{\vb}{\mathbf{v}}
\newcommand{\RR}{\mathds{R}}

\newcommand{\zero}{\mathbf{0}}

\newcommand{\vecz}[1]{\mathrm{vec}\left(#1 \right)}

\newcommand{\inner}[2]{\left\langle #1, #2 \right\rangle}

\newcommand{\norm}[1]{\left\| #1 \right\|}

\begin{document}
\twocolumn[

\aistatstitle{Characterization of Gradient Dominance and Regularity Conditions for Neural Networks}

\aistatsauthor{ Yi Zhou \And Yingbin Liang}

\aistatsaddress{Ohio State University} ]

\begin{abstract}
The past decade has witnessed a successful application of deep learning to solving many challenging problems in machine learning and artificial intelligence. However, the loss functions of deep neural networks (especially nonlinear networks) are still far from being well understood from a theoretical aspect. In this paper, we enrich the current understanding of the landscape of the square loss functions for three types of neural networks. Specifically, when the parameter matrices are square, we provide explicit characterization of the global minimizers for linear networks, linear residual networks, and nonlinear networks with one hidden layer. Then, we establish two quadratic types of landscape properties for the square loss of these neural networks, i.e., the gradient dominance condition within the neighborhood of their full rank global minimizers, and the regularity condition along certain directions and within the neighborhood of their global minimizers. These two landscape properties are desirable for the optimization around the global minimizers of the loss function for these neural networks.   
\end{abstract}

\section{Introduction}\label{sec:introduction}

The significant success of deep learning (see, e.g., \cite{deeplearning_book}) has influenced many fields such as machine learning, artificial intelligence, computer vision, natural language processing, etc.  Consequently, there is a rising interest in understanding the fundamental properties of deep neural networks. Among them, the landscape (also referred to as geometry) of the loss functions of neural networks is an important aspect, since it is central to determine the performance of optimization algorithms that are designed to minimize these functions. The loss functions of neural networks are typically nonconvex, and hence understanding these functions requires significantly new insights and analysis techniques.

There has been a growing literature recently that contributed towards understanding the landscape properties of loss functions of neural networks. For example, \cite{Baldi_1989_1} showed that any local minimum is a global minimum for the square loss function for linear networks with one hidden layer, and more recently \cite{Kawaguchi_2016,Yun_2017} showed that such a result continues to hold for deep linear networks. \cite{Choromanska_2015_1,Choromanska_2015_2} 
characterized the distribution properties of the local minimizers for deep nonlinear networks, and \cite{Kawaguchi_2016} further eliminated some assumptions in \cite{Choromanska_2015_1}, and established the equivalence between the local minimum and the global minimum.
More results on this topic are further discussed in \Cref{sec:relatedwork}.

The main focus of this paper is on two important landscape properties that have been shown to be important to determine the convergence of the first-order algorithms for nonconvex optimization. The first property is referred to as {\em gradient dominance condition} as we describe below. Consider a global minimizer $\xb^*$ of a generic function $f: \RR^d \to \RR$, and a neighborhood $\mathcal{B}_{\xb^*}(\delta)$ around $\xb^*$. The (local) gradient dominance condition with regard to $\xb^*$ is given by
\begin{align*}
\forall \xb \in \mathcal{B}_{\xb^*}(\delta),~ f(\xb) - f(\xb^*) \le \lambda \norm{\nabla f(\xb)}_2^2,
\end{align*}
where $\lambda > 0$ and $\mathcal{B}_{\xb^*}(\delta)$ is a neighborhood of $\xb^*$. This condition is a special case of the {\L}ojasiwiecz gradient inequality \cite{Lojasiewicz_book} (with exponent $\theta = 2$), and has been shown to hold for a variety of machine learning problems, e.g., a square loss function for phase retrieval \cite{Zhou_2016} and blind deconvolution \cite{Xiaodong_2016}. If the algorithm iterates in the neighborhood $\mathcal{B}_{\xb^*}(\delta)$, then the gradient dominance condition, together with a Lipschitz property of the gradient of an objective function, guarantees a linear convergence of the function value residual $f(\xb) - f(\xb^*)$ \cite{Karimi_2016,Reddi_2016}.  

The second property is referred to as {\em regularity condition}, with the (local) regularity condition given by
\begin{align*}
&\forall \xb \in \mathcal{B}_{\xb^*}(\delta), \\
&\inner{\xb - \xb^*}{\nabla f(\xb)} \ge \alpha \norm{\nabla f(\xb)}_2^2 + \beta \norm{\xb - \xb^*}_2^2,
\end{align*}
where $\alpha, \beta > 0$. This condition can be viewed as a restricted version of the strong convexity, and it has been shown to guarantee a linear convergence of the iterate residual $\norm{\xb - \xb^*}$ in this local neighborhood \cite{Nesterov_2014,Candes_2014}. Problems such as phase retrieval \cite{Candes_2014}, affine rank minimization \cite{Zheng_2015,tu2015low} and matrix completion \cite{zheng2016convergence} have been shown to satisfy the local regularity condition. 

However, these two properties have not been explored thoroughly for the loss functions of neural networks with only very few exceptions. \cite{Moritz_2016} established the gradient dominance condition for a linear residual network (in which each residual unit having only one linear layer) within a local neighborhood of the origin. The goal of this paper is to explore these two geometric conditions for a much broader types of neural networks. In particular, we focus on three types of neural networks: feed forward linear neural networks \cite{Lippmann_1988}, linear residual neural networks \cite{Kaiming_resnet}, and nonlinear neural networks with one hidden layer. 

\subsection{Our Contributions}

We study the square loss function of linear, linear residual, and one-hidden-layer nonlinear neural networks. We focus on  the scenario, in which all parameter matrices of the neural networks are square so that the global minimizers, gradient and Hessian of loss functions can be expressed in a trackable form for analysis . We first characterize the form of global minimizers of these loss functions, and then establish local gradient dominance and regularity conditions for these loss functions.

\textbf{Characterization of global minimizers:} For deep linear neural networks, we show that global minimizers can be uniquely characterized in an explicit form up to an equivalence class. Furthermore, all the global minimizers correspond to parameter matrices that are full rank. We then extend such a result to further characterize the full-rank global minimizers of deep linear residual networks and one-hidden-layer nonlinear neural networks. Our results generalize the characterization of global minimizers of {\em shallow} linear networks in \cite{Baldi_1989_1} to deep linear, residual and one-hidden-layer nonlinear neural networks.

\textbf{Gradient dominance condition:} For deep linear networks, we show that the gradient dominance condition holds within the neighborhood of any global minimizer, and hence any critical point within such a neighborhood is also a global minimizer. We further show that the same result also holds in parallel for deep linear residual networks within the neighborhood of any full-rank global minimizer, and for nonlinear networks with one hidden layer within the neighborhood of any global minimizer. Moreover, comparing the gradient dominance condition of the two types of linear networks, the identity shortcut in the residual networks helps to regularize the constant of the gradient dominance condition in the neighborhood of the origin to be more amenable for optimization. Our results generalize that in \cite{Moritz_2016} within the neighborhood of the origin for residual networks with shortcut depth $r=1$ to the neighborhood of any full-rank global minimizer for residual networks with $r >1$.

\textbf{Regularity condition:} For deep linear networks, we establish the local regularity condition within the neighborhood of any global minimizer along certain directions. We further show that the same result also holds in parallel for deep linear residual networks and one-hidden-layer nonlinear neural networks. Comparing the local regularity condition of the two types of linear networks, the identity shortcut in residual networks broadens the range of directions along which the regularity condition holds in the neighborhood of the origin. Hence, the global minimizers of the linear residual networks near the origin open a larger aperture of attraction for optimization paths than that of the global minimizer of the linear networks.



\subsection{Related Work}\label{sec:relatedwork}
\textbf{Gradient dominance condition and regularity condition for nonconvex problems}: 
As we discussed above, the gradient dominance condition have recently been exploited to characterize the linear convergence of first-order algorithms for nonconvex optimization \cite{Karimi_2016,Reddi_2016}. This condition was established for problems such as phase retrieval \cite{Zhou_2016}, blind deconvolution \cite{Xiaodong_2016}, and linear residual neural networks \cite{Moritz_2016}.  The regularity condition has also been exploited to characterize the linear convergence of first-order algorithms for nonconvex optimization \cite{Candes_2014}.  This condition was established for phase retrieval \cite{Candes_2014,chen2015solving,Huishuai_2016,wang2016solving}, for affine rank minimization \cite{Zheng_2015,tu2015low,white2015local}, and for matrix completion problems \cite{chen2015fast,zheng2016convergence}.


\textbf{Other landscape properties of linear networks}:
The study of the landscape of the square loss function for linear neural networks dates back to the pioneering work \cite{Baldi_1989_1,Baldi_1989_2}. There, they studied the autoencoder with one hidden layer and showed the equivalence between the local minimum and the global minimum with a characterization of the form of global minimum points. \cite{Baldi_2012} further generalizes these results to the complex-valued autoencoder setting. The equivalence between the local minimum and the global minimum of deep linear networks was established in \cite{Kawaguchi_2016,Lu_2017,Yun_2017} respectively under different assumptions. In particular, \cite{Yun_2017} established the necessary and sufficient conditions for a critical point of the deep linear network to be a global minimum. The same result was shown in \cite{Freeman_2017} for deep linear networks with the width of intermediate layers being larger than the input and output layers. \cite{Taghvaei_2017} studied the effect of regularization on the critical points for a two-layer linear network. 
\cite{Sihan_2016} studied the property of the Hessian matrix for deep linear residual networks.

\textbf{Other landscape properties of nonlinear networks}:
There have also been studies on understanding the landscape of nonlinear neural networks from theoretical perspectives.
\cite{Yu_1995} considered a one-hidden-layer nonlinear neural network with sigmoid activation and showed that all local minimum are also global minimum provided that the number of input units equals the number of data samples. \cite{Gori_1992} studied a class of multi-layer nonlinear neural networks, and showed that all critical points of full column rank achieve the global minimum with zero loss, if the sample size is less than the input dimension and the widths of the layers form a pyramidal structure. \cite{Nguyen_2017} further generalized the results in \cite{Gori_1992} to a larger class of nonlinear networks and showed that critical points with non-degenerate Hessian are the global minimum. \cite{Choromanska_2015_1,Choromanska_2015_2} connected the loss function of deep nonlinear networks with the Hamiltonian of the spin-glass model under certain assumptions and characterized the distribution properties of the local minimizers. Then, \cite{Kawaguchi_2016} further eliminated some of the assumptions in \cite{Choromanska_2015_1}, and established the equivalence between the local minimum and the global minimum by reducing the loss function of the deep nonlinear network to that of the deep linear network. \cite{Soltanolkotabi_2017} established the local strong convexity of overparameterized nonlinear networks with one hidden layer and quadratic activation functions. Furthermore, \cite{Zhong_2017}  established the local strong convexity of a class of nonlinear networks with one hidden layer with the Gaussian input data, and established the local linear convergence of gradient descent method with tensor initialization. 
\cite{Soudry_2017} studied a one-hidden-layer nonlinear neural network with piecewise linear
activation function and a single output, and showed that the volume of differentiable regions of the empirical
loss containing sub-optimal differentiable local minima is exponentially vanishing
in comparison with the same volume of global minima as the number of data samples tends to infinity. 
\cite{Xiebo_2016} studied the nonlinear neural network with one hidden layer, and showed that a diversified weight can lead to good generalization error. \cite{Dauphin_2014} investigated the saddle point issues in deep neural networks, motivated by the results from statistical physics and random matrix theory. 
Recently, \cite{Feizi_2015} studied a one-hidden-layer nonlinear neural network with the parameters constrained in a finite set of lines, and showed that most local optima are global optima.

\section{Preliminaries of Three Neural Networks}\label{sec:models}
In this section, we describe the square loss functions that we consider for three types of neural networks, and characterize the forms of global minimizers of these loss functions, which further help to establish our main results of landscape properties for these loss functions in \Cref{sec: gdcond,sec: regcond}. 

Throughout, $(\Xb, \Yb)$ denotes the input and output data matrix pair. We assume that $\Xb, \Yb \in \RR^{d \times m}$, i.e., there are $m$ data samples. We denote $\Sigmab_{\Yb\Yb}:=\Yb\Yb^\top$. We assume that $\Sigmab_{\Xb\Xb} := \Xb\Xb^\top$ and $\Sigmab_{\Xb\Yb}:=\Xb\Yb^\top$ are full rank, and assume that $\Sigmab:=\Sigmab_{\Xb\Yb}^\top \Sigmab_{\Xb\Xb}^{-1} \Sigmab_{\Xb\Yb}$ has distinct eigenvalues. Note that these are standard assumptions adopted in \cite{Baldi_1989_1,Kawaguchi_2016}.

We also adopt the following notations. For a matrix $\Xb\in \RR^{d\times m}$, we denote $\vecz{\Xb}\in \RR^{dm\times 1}$ as the vertical stack of the columns of $\Xb$, i.e., $\vecz{\Xb}:=[\xb_1^\top \; \xb_2^\top \; \ldots\; \xb_m^\top]^\top$. The kronecker product between matrices $\Xb$ and $\Yb$ is denoted as $\Xb \otimes \Yb$.  For a matrix $\Xb$, the spectral norm is denoted by $\|\Xb\|$, the smallest nonzero singular value is denoted by $\eta_{\min}(\Xb)$, and the trace is denoted by $\mathrm{Tr}(\Xb)$. We also denote a collection of natural numbers as $[n]:=\{1,\ldots,n\}$. 

\subsection{Linear Neural Networks}
Consider a feed forward linear neural network with $l-1$ hidden layers. Each layer $k\in[l]$ is parameterized by a matrix $\Wb_k$, and we use $\Wb:=\{\Wb_1,\ldots, \Wb_l \}$ to denote the collection of all model parameter matrices. In particular, we consider the setting where all parameter matrices are square, i.e., $\Wb_k \in \RR^{d\times d}$ for all $k\in [l]$. We are interested in understanding the properties of the following square loss function in training the network as adopted by \cite{Baldi_1989_1,Kawaguchi_2016}:
\begin{align}\label{eq:linearloss}
h(\Wb) := \tfrac{1}{2} \norm{\Wb_l\Wb_{l-1}\ldots\Wb_1 \Xb - \Yb}_F^2,
\end{align}
where $\norm{\cdot}_F$ denotes the matrix Frobenius norm. 

It can be observed that the set of the global minimizers of $h(\Wb)$ is invariant under invertible transformations. Namely, if $\{\Wb_l^*,\ldots, \Wb_1^* \}$ is a global minimizer, then $\{\Wb_l^*\Cb_l,\Cb_l^{-1}\Wb^*_{l-1}\Cb_{l-1},\ldots,\Cb_2^{-1}\Wb_1^*\}$ is also a global minimizer, where $\Cb_2,\ldots, \Cb_l$ are arbitrary invertible square matrices. Thus, we treat all global minimizers up to such invertible matrix transformation as an {\em equivalence class}. The following result states that under certain conditions, the global minimizers of $h(\Wb)$ can be uniquely characterized up to an equivalent class.
\begin{proposition}\label{thm: optima_linear}
Consider $h(\Wb)$ of a linear neural network with square parameter matrices. Then the global minimizers can be uniquely (up to an equivalence class) characterized by
		\begin{align}
		\Wb_l^* &= \Ub \Cb_l, \nonumber\\
		&\cdots \nonumber\\
		\Wb_{k}^* &= \Cb_{k+1}^{-1}\Cb_k, \nonumber\\
		\Wb_1^* &= \Cb_2^{-1}\Ub^\top \Sigmab_{\Xb\Yb}^\top \Sigmab_{\Xb\Xb}^{-1}, \label{eq:optlinearloss}
		\end{align}
		where $\Cb_2,\ldots, \Cb_l\in \RR^{d\times d}$ are arbitrary invertible matrices and $\Ub = [\ub_1,\ldots, \ub_d]$ is the matrix formed by the  eigenvectors corresponding to the top $d$ eigenvalues of $\Sigmab$. In such a case, the global minimal value of $h(\Wb)$ is given by $\mathrm{Tr}(\Sigmab_{\Yb\Yb})-\sum_{i=1}^{d} \lambda_i$,	where $\{\lambda_i, i\in [d]\}$ are the top $d$ eigenvalues of $\Sigmab$.
		
\end{proposition}

\Cref{thm: optima_linear} generalizes the characterization of the global minimizers of shallow linear networks in \cite{Baldi_1989_1} to deep linear networks. It states a not-so-obvious fact that any global minimizer in such a case {\em must} take the form in \cref{eq:optlinearloss}, although it is easy to observe that $\Wb^*$ given in \cref{eq:optlinearloss} achieves the global minimum. Moreover, the following corollary follows as an immediate observation from \cref{eq:optlinearloss}. 
\begin{coro}\label{cor:fullrank}
Any global minimizer of $h(\Wb)$ of a linear neural network with square parameter matrices must be full rank.
\end{coro}

\subsection{Linear Residual Neural Networks}

Consider the linear residual neural network, which further introduces the residual structure to the linear neural network. That is, one adds a shortcut (identity map) for every, say, $r$ hidden layers. Assuming we have in total $l$ residual units, then we consider the square loss of a linear residual neural network given as follows:
\begin{align}\label{eq:residualloss}
f(\Ab) := \tfrac{1}{2} \|(\Ib+&\Ab_{lr}\ldots\Ab_{l1})\ldots(\overbrace{\Ib+\Ab_{kr}\ldots\Ab_{k1}}^{\Wb_{k}})\nonumber\\
&\cdots(\Ib+\Ab_{1r}\ldots\Ab_{11}) \Xb - \Yb\|_F^2,
\end{align}
where the model parameters of each layer are denoted by $\Ab_{kq}, k\in [l], q\in [r]$. Again, we consider the case when all parameter matrices are square, i.e., $\Ab_{kq} \in \RR^{d\times d}$ for all $k\in [l], q\in [r]$. The following property of the linear residual neural network follows directly from \Cref{thm: optima_linear} for the linear nuerual network.
\begin{proposition}\label{thm: form_resnet}
Consider $f(\Ab)$ of the linear residual network with $m = d$. Then a full-rank global minimizer $\Ab^*$ is fully characterized as, for all $k\in [l]$
	\begin{align}\label{eq:amatrix}
	\Ab_{kr}^* &= \Ub_k \Cb_{kr}, \nonumber\\
	&\cdots \nonumber\\
	\Ab_{kq}^* &= \Cb_{k(q+1)}^{-1}\Cb_{kq}, \nonumber\\
	&\cdots \nonumber\\
	\Ab_{k1}^* &= \Cb_{k2}^{-1}\Ub_k^\top (\Wb_k^* - \Ib),
	\end{align}
	where $W_k^*$ for $k\in [l]$ is characterized as
	\begin{align}\label{eq:wmatrix}
	\Wb_l^* &= \widehat{\Ub} \widehat{\Cb}_l, \nonumber\\
	&\cdots \nonumber\\
	\Wb_{k}^* &= \widehat{\Cb}_{k+1}^{-1}\widehat{\Cb}_k, \nonumber\\
	&\cdots \nonumber\\
	\Wb_1^* &= \widehat{\Cb}_2^{-1}\widehat{\Ub}^\top \Sigmab_{\Xb\Yb}^\top\Sigmab_{\Xb\Xb}^{-1}.
	\end{align}
	Here, $\Cb_{k2},\ldots, \Cb_{kl}$ for all $k\in [l]$ and $\widehat{\Cb_2},\ldots, \widehat{\Cb_l}$ are arbitrary invertible matrices, and $\Ub_k = [\ub_{k1},\ldots, \ub_{kd}]$ is the matrix formed by all the eigenvectors of $(\Wb_k^* - \Ib)(\Wb_k^* - \Ib)^\top$ and $\widehat{\Ub} = [\hat{\ub}_1,\ldots, \hat{\ub}_d]$ is the matrix formed by all the eigenvectors of $\Sigmab$.
	
\end{proposition}

The above result characterizes the full-rank global minimizers $\Ab^*$ via the form given by \cref{eq:amatrix}. In particular, the characterization in \cref{eq:wmatrix} implies that all residual units are also full rank. We note that when $r=1$, the above characterization is consistent with the construction of the global minimizer in \cite{Moritz_2016}.

\subsection{Nonlinear Neural Network with One-hidden-layer}
Consider a nonlinear neural network with one hidden layer, where the layer parameters are square, i.e., $\Wb_1 \in \RR^{d\times d}$ and $\Wb_2 \in \RR^{d\times d}$, and each hidden neuron adopts a differentiable nonlinear activation function $\sigma: \RR \to \RR$. We consider the following square loss function
\begin{align}\label{eq: nonlinear}
g(\Wb) := \tfrac{1}{2} \norm{\Wb_2 \sigma(\Wb_{1} \Xb) - \Yb}_F^2,
\end{align}
where $\sigma$ acts on $\Wb_1\Xb$ {\em entrywise}. In particular, we consider a class of activation functions that satisfy the condition $\mathrm{range}(\sigma) = \RR$. A typical example of such activation function is the class of parametric ReLU activation functions, i.e., $\sigma(x) = \max \{x, ax\}$, where $0<a<1$. 

The following result characterizes the form of the global minimizers of $g(\Wb)$.
\begin{proposition}\label{prop: globmin_nonlinear}
Consider $g(\Wb)$ of the one-hidden-layer nonlinear neural networks with $m = d$ and $\mathrm{range}(\sigma) = \RR$. Then any global minimizer $\Wb^*$ can be characterized as 
	\begin{align}\label{eq: nonlinear_globalmin}
	\Wb_2^* = \widetilde{\Wb}_2^*, ~\Wb_1^* = \sigma^{-1}(\widetilde{\Wb}_1^* \Xb)\Xb^{-1},
	\end{align}
	where $(\widetilde{\Wb}_2^*, \widetilde{\Wb}_1^*)$ is a global minimizer of the corresponding linear network, and is fully characterized by \Cref{thm: optima_linear}.
\end{proposition}
We note that the inverse function should be understood as $\sigma^{-1}(y) := \{x:\sigma(x) = y \}$, and is well defined since $\mathrm{range}(\sigma) = \RR$.
It can be seen from \cref{eq: nonlinear_globalmin} that the global minimizers of $g(\Wb)$ satisfy $\sigma(\Wb_{1}^* \Xb) = \widetilde{\Wb}_1^* \Xb$, which is {\em full rank} by \Cref{cor:fullrank}.

Based on the results in this section, we observe that for the scenario where all parameter matrices are square, all/partial global minimizers of all three aforestudied neural networks consist of full rank parameter matrices. Such a property further assists the establishment of the gradient dominance condition and the regularity condition for the loss functions of these types of neural networks, as we present in \Cref{sec: gdcond,sec: regcond}.

\section{Gradient Dominance Condition}\label{sec: gdcond}

The gradient dominance condition is generally regarded as a useful property that can be exploited for analyzing the convergence performance of optimization methods. In particular, this condition, together with a Lipschitz property of the gradient of an objective function, guarantees the linear convergence of the function value sequence generated by the gradient descent method. In this section, we establish the gradient dominance condition for the three types of neural networks of interest in this paper.

\subsection{Linear Neural Networks}
For the linear network, we define $\vecz{\Wb}: = [\vecz{\Wb_1}^\top\; \vecz{\Wb_2}^\top \; \ldots \; \vecz{\Wb_l}^\top]^\top$, and denote $\eb(\Wb):= \Wb_l\Wb_{l-1}\ldots\Wb_1 \Xb - \Yb$ as the error matrix.
We start our analysis by exploring the gradient of $h(\Wb)$. We present them as follows in the denominator layout (i.e., column layout), and all calculations are provided in the supplemental material for the reader's convenience.

The $k$-th block of the gradient of $h(\Wb)$ can be characterized as
\begin{align}
\forall k \in[l], \quad \nabla_{\vecz{\Wb_k}} h(\Wb)  = \Gb_k(\Wb)^\top \vecz{\eb},\label{eq: grad_linear}
\end{align}
where $\eb=\eb(\Wb)$ is the error matrix of $h(\Wb)$, and
\begin{align}\label{eq: G_k}
\Gb_k(\Wb) := (\Wb_{k-1}\ldots\Wb_1\Xb)^\top\otimes (\Wb_l\ldots\Wb_{k+1}).
\end{align}
Note that for the boundary cases $k\in \{1,l\}$,  $\Wb_0\Wb_1$ and $\Wb_{l}\Wb_{l+1}$ should be understood as identity matrix.  

We next establish the gradient dominance condition in the neighborhood of any global minimizer of $h(\Wb)$ for the linear networks.
\begin{thm}\label{thm: gd_condition}
	Consider $h(\Wb)$ of the linear neural network with $m=d$. Consider a global minimizer $\Wb^*$ and let $\tau= \tfrac{1}{2}\min_{k\in[l]} \eta_{\min}(\Wb_k^*) $. Then any point in the neighborhood of $\Wb^*$ defined as $\{\Wb:\|\Wb_k-\Wb_k^*\|<\tau,~ \forall k\in [l]\}$ satisfies
	\begin{align}\label{eq:gd_linear}
	h(\Wb) - h(\Wb^*) \le \lambda_h \norm{\nabla_{\vecz{\Wb}} h(\Wb)}_2^2,
	\end{align}
	where $\lambda_h = (2l\tau^{2(l-1)}\eta_{\min}^2(\Xb))^{-1}$.
	Consequently, any critical point in this neighborhood is a global minimizer.
\end{thm}
We note that \Cref{cor:fullrank} guarantees that any global minimizer $\Wb^*$ of $h(\Wb)$ under the assumption of \Cref{thm: gd_condition} is full rank, and hence the parameter $\tau$ defined in \Cref{thm: gd_condition} is strictly positive. The gradient dominance condition implies a linear convergence of the function value to the global minimum via a gradient descent algorithm, if the iterations of the algorithm stay in this $\tau$ neighborhood. In particular, a larger parameter $\tau$ (a larger minimum singular value) implies a smaller $\lambda_h$, which yields a faster convergence of the function value to the global minimum via a gradient descent algorithm. 


\subsection{Linear Residual Neural Networks}
For the linear residual network, we define $\vecz{\Ab}: = [\vecz{\Ab_{11}}^\top\; \vecz{\Ab_{12}}^\top\; \ldots \; \vecz{\Ab_{lr}}^\top]^\top$. For all $k\in [l]$, we denote $\Wb_k := \Ib + \Ab_{kr}\ldots\Ab_{k1}$, and denote $\eb(\Wb):= \Wb_l\Wb_{l-1}\ldots\Wb_1 \Xb - \Yb$ as the error matrix. Please note that $\Wb_k, k\in [l]$ here are only notations for convenience, and the real parametrization is by the matrices $\Ab_{kq}, k\in [l], q\in [r]$.

We derive the following first-order derivatives of $f(\Ab)$ with details given in the supplemental material. The $(k,q)$-th block of the first-order derivative of $f(\Ab)$ can be characterized as
\begin{align}
\forall k\in [l], &q\in [r], \nonumber\\
&\nabla_{\vecz{\Ab_{kq}}} f(\Ab)  = \Qb_{kq}(\Ab)^\top \vecz{\eb}, \label{eq: grad_resnet}
\end{align}
where matrix $\Qb_{kq}$ takes the form
\begin{align}\label{eq: Q_kq}
\Qb_{kq}&(\Ab) := \left[\left(\Wb_{k-1}\ldots\Wb_1\Xb \right)^\top \otimes  \left(\Wb_l\ldots\Wb_{k+1} \right)\right] \nonumber\\
&\left[\left(\Ab_{k(q-1)}\ldots\Ab_{k1}\right)^\top \otimes  \left(\Ab_{kr}\ldots\Ab_{k(q+1)}  \right) \right].
\end{align}

We then obtain the following gradient dominance condition within the neighborhood of a full-rank global minimizer.
\begin{thm}\label{thm: gd_resnet_1}
	Consider $f(\Ab)$ of the linear residual neural network with $m=d$.  Consider a full-rank global minimizer $\Ab^*$, and let $\tau= \tfrac{1}{2}\min_{k\in[l]} \eta_{\min}(\Wb_k^*)$, $ \tilde{\tau}= \tfrac{1}{2}\min_{k\in[l],q\in[r]} \eta_{\min}(\Ab_{kq}^*),$ and pick $\hat{\tau}$ sufficiently small such that any point in the neighborhood of $\Ab^*$ defined as $\{\Ab: \|\Ab_{kq}-\Ab_{kq}^*\|<\hat{\tau}, \forall k\in [l], q\in [r]\}$ satisfies $\|\Wb_{k}-\Wb_{k}^*\|<\tau$ for all $k\in[l]$. Then any point in the neighborhood of $\Ab^*$ defined as $\{\Ab: \|\Ab_{kq}-\Ab_{kq}^*\|<\min\{\hat{\tau}, \tilde{\tau}\}, \forall k\in [l], q\in [r]\}$ satisfies
	\begin{align}\label{eq:gd_resnet}
	f(\Ab) - f(\Ab^*) \le \lambda_{f} \norm{\nabla_{\vecz{\Ab}} f(\Ab)}_2^2,
	\end{align}
	where $\lambda_{f} = \left(2lr\tilde{\tau}^{2(r-1)} \tau^{2(l-1)} \eta_{\min}^2(\Xb) \right)^{-1}.$ Consequently, any critical point in this neighborhood is a global minimizer.
\end{thm}

We note that the above theorem establishes the gradient dominance condition around full rank global minimizers. This is not too restrictive as $\Wb_k^*$ is guaranteed to be full rank for all $k$ by \cref{eq:wmatrix}. Then, $\Ab^*$ is full rank if and only if $\Wb_k^* - \Ib$ is full rank, which is satisfied if the column space of $\Wb_k^*$ is incoherent with its row space. 
Also, positive $\hat{\tau}$ exists because by continuity $\|\Wb_{k}-\Wb_{k}^*\| \to 0$ as $\|\Ab_{kq}-\Ab_{kq}^*\|\to 0$ for all $q\in[r]$. 

As a comparison to the gradient dominance condition obtained in \cite{Moritz_2016}, which is applicable to the neighborhood of the origin for the residual network with $r=1$, the above result characterizes the gradient dominance condition for a broader range of the parameter space, which is applicable to the neighborhood of any full-rank minimizer $\Ab^*$ and to more general residual networks with $r>1$.

We note that the parameter $\lambda_{f}$ in \Cref{thm: gd_resnet_1} depends on both $\tau$ and $\tilde{\tau}$, where $\tau$ captures the overall property of each residual unit and $\tilde{\tau}$ captures the property of individual linear unit in each residual unit. Hence, in general, the $\lambda_{f}$ in \Cref{thm: gd_resnet_1} for linear residual networks is very different from the $\lambda_{h}$ in the gradient dominance condition in \Cref{thm: gd_condition} for linear networks. When the shortcut depth $r$ becomes large,  the parameter $\lambda_{f}$ involves $\tilde{\tau}$ that depends on more unparameterized variables in $\Ab^*$, and hence becomes more similar to the parameter $\lambda_h$ of linear networks.

To further compare the $\lambda_{f}$ in \Cref{thm: gd_resnet_1} and the $\lambda_{h}$ in \Cref{thm: gd_condition}, consider a simplified setting of the linear residual network with the shortcut depth $r=1$. Then $\lambda_{f} = \left(2l \tau^{2(l-1)} \eta_{\min}^2(\Xb) \right)^{-1}$, which is better regularized, although it takes the same expression as $\lambda_{h}$ in \Cref{thm: gd_condition} for the linear network. The reason is that for residual networks, all $\Wb_k, k\in[l]$ are further parameterized  by $\Ib + \Ab_k$. When $\norm{\Ab_k}$ is small (in particular, less than one), $\eta_{\min} (\Ib + \Ab_k)$ (and hence the parameter $\tau$) is regularized away from zero by the identity map, which was also observed by \cite{Moritz_2016}. Consequently, the identity shortcut leads to a smaller $\lambda_f$ (due to larger $\tau$) compared to a large $\lambda_h$ when the parameters of linear networks have small spectral norm. Such a smaller $\lambda_f$ is more desirable for optimization, because the function value approaches closer to the global minimum after one iteration of a gradient descent algorithm.



\subsection{Nonlinear Neural Network with One Hidden layer}
For a nonlinear network with one hidden layer, we define $\vecz{\Wb}: = [\vecz{\Wb_1}^\top\; \vecz{\Wb_2}^\top]^\top$ and denote $\eb(\Wb):= \Wb_2 \sigma(\Wb_{1} \Xb) - \Yb$ as the error matrix.

For such nonlinear networks, the structures of the gradient and Hessian of the loss function $g(\Wb)$ are much more complicated than those of linear networks. In specific, the gradient of $g(\Wb)$ can be characterized as follows (see supplemental material for the derivation)
\begin{align}
\nabla_{\vecz{\Wb_2}} g(\Wb) &= (\sigma(\Wb_1\Xb) \otimes \Ib) \vecz{\eb}, \label{eq: nonlinear_grad_1}\\
\nabla_{\vecz{\Wb_1}} g(\Wb) &= (\Xb \otimes \Ib)\vecz{\sigma'(\Wb_1\Xb) \circ (\Wb_2^\top \eb)}, \label{eq: nonlinear_grad_2}
\end{align}
where ``$\circ$'' denotes the entrywise Hadamard product, and $\sigma'(\cdot)$ denotes the derivative of $\sigma(\cdot)$.
The following result establishes the gradient dominance condition for one-hidden-layer nonlinear neural networks. 
\begin{thm}\label{thm: gd_nonlinear}
Consider the loss function $g(\Wb)$ of one-hidden-layer nonlinear neural networks with $m=d$ and with $\mathrm{range}(\sigma) = \RR$. Consider a global minimizer $\Wb^*$, and let $\tau = \frac{1}{2} \eta_{\min} (\sigma(\Wb_1^*\Xb))$. Then any point in the neighborhood of $\Wb^*$ defined as $\{\Wb: \norm{\sigma(\Wb_1\Xb) - \sigma(\Wb_1^*\Xb)}\le \tau \}$ satisfies
	\begin{align}
	g(\Wb) \le \lambda_g \norm{\nabla_{\vecz{\Wb}} g(\Wb)}_2^2,
	\end{align}
	where $\lambda_g = (2\tau^2)^{-1}$. 
\end{thm}
We note that the characterization in \Cref{prop: globmin_nonlinear} guarantees that $\sigma(\Wb_1^*\Xb)$ is full rank, and hence $\tau$ is well defined. Differently from linear networks, the gradient dominance condition for nonlinear networks holds in a nonlinear $\tau$ neighborhood that involves the activation function $\sigma$. This is naturally due to the nonlinearity of the network. Furthermore, the parameter $\tau$ depends on the nonlinear term $\sigma(\Wb_1\Xb)$, whereas the $\tau$ in \Cref{thm: gd_condition} of linear networks depends on the individual parameters $\Wb_k$.


\section{Regularity Condition}\label{sec: regcond}

The regularity condition is an important landscape property in optimization theory, which has been shown to guarantee the linear convergence of a gradient descent iteration sequence to a global minimizer in various nonconvex problems as we discuss in \Cref{sec:introduction}. In this section, we establish the regularity condition for the loss functions of the three neural networks of interest here.
\subsection{Linear Neural Networks}
Consider any global minimizer $\Wb^*$ of $h(\Wb)$. We focus on the more amenable case where $m=d$. In such a case, the global minimal value is zero (i.e., $h(\Wb^*) = 0$), and the Hessian at the global minimizers can be expressed in a trackable form. More specifically, the $(k,k')$-th block of the Hessian of $h(\Wb^*)$ can be characterized as
\begin{align}
&\forall k, k' \in [l], \nonumber\\
&\nabla_{\vecz{\Wb_k'}} \big(\nabla_{\vecz{\Wb_k}} h(\Wb^*) \big) = \Gb_{k'}(\Wb^*)^\top\Gb_k(\Wb^*), \label{eq: hessian_linear}
\end{align}
where the matrix $\Gb_k$ is given in \cref{eq: G_k}.
By further denoting $\Gb(\Wb):= [\Gb_1(\Wb), \ldots, \Gb_l(\Wb)]$, the entire Hessian matrix at any global minimizer $\Wb^*$ can be written as
\begin{align}
\nabla_{\vecz{\Wb}}^2 h(\Wb^*) = \Gb(\Wb^*)^\top \Gb(\Wb^*).
\end{align}

The following result establishes the regularity condition for the square loss function for linear networks.
\begin{thm}\label{thm: rc_condition}
Consider $h(\Wb)$ of linear neural networks with $m=d$. Further consider a global minimizer $\Wb^*$, and let $\zeta = 2\max_{k\in[l]} \norm{\Wb_k^*}$. Then for any $\delta>0$, there exists a sufficiently small $\epsilon (\delta)$ such that any point $\Wb$ that satisfies 
\begin{equation}\label{eq:dircond}
\norm{\Gb(\Wb^*) \vecz{\Wb- \Wb^*}}_2 \ge \delta \norm{\vecz{\Wb- \Wb^*}}_2
\end{equation}
	and within the neighborhood of $\Wb^*$ defined as $\{\Wb: \|\Wb_k-\Wb_k^*\|_F<\epsilon(\delta), \forall k\in[l] \}$ satisfies
	\begin{align}
	&\langle\nabla_{\vecz{\Wb}} h(\Wb), \vecz{\Wb- \Wb^*} \rangle\nonumber\\
	&\ge \alpha \norm{\nabla_{\vecz{\Wb}} h(\Wb)}_2^2 + \beta \norm{\vecz{\Wb- \Wb^*}}_2^2, \label{eq: regular}
	\end{align}
	where $\alpha = \gamma/(l\zeta^{2(l-1)} \norm{\Xb}^2)$ and $\beta = (1-\gamma) \delta^2/2$ for any $\gamma\in (0,1)$.
\end{thm}

We note that the regularity condition as in \cref{eq: regular} has been established and exploited for the convergence analysis in various nonconvex problems such as phase retrieval and rank minimization (see the references in Section \ref{sec:relatedwork}). There, the regularity condition was shown to hold within the entire neighborhood of any global minimizer. In comparison, \Cref{thm: rc_condition} guarantees the regularity condition for linear neural networks within a neighborhood of $\Wb^*$ with the further constraint in \cref{eq:dircond}. It can be observed that the condition in \cref{eq:dircond} does not depend on the norm of $\vecz{\Wb-\Wb^*}$, and hence it is a condition only on the direction of $\vecz{\Wb-\Wb^*}$, along which the regularity condition can be satisfied. Furthermore, the parameter $\delta$ in \cref{eq:dircond} determines the range of directions that satisfy \cref{eq:dircond}. For example, if we set $\delta = \eta_{\min}(\Gb(\Wb^*))$, then all $\Wb$ such that $\vecz{\Wb- \Wb^*} \perp \ker{\Gb(\Wb^*)}$ satisfy the condition \cref{eq:dircond}. 

For all $\Wb$ with directions that satisfy \cref{eq:dircond} and hence satisfy the regularity condition, it can be shown that one gradient descent iteration yields an update that is closer to the global minimizer $\Wb^*$ \cite{Candes_2014}. Hence, $\Wb^*$ serves as an attractive point along those directions of $\Wb$ that satisfy \cref{eq:dircond}.  Furthermore, the value of $\delta$ in \cref{eq:dircond} affects the parameter $\beta$ in the regularity condition. Larger $\delta$ results larger $\beta$, which further implies that one gradient descent iteration at the point $\Wb$ yields an update that is closer to the global minimizer $\Wb^*$ \cite{Candes_2014}.

\subsection{Linear Residual Neural Networks}
Consider any global minimizer $\Ab^*$ of $f(\Ab)$ of a linear residual network. Suppose, $m=d$, which implies $f(\Ab^*)=0$. Then the $(k,q)-(k',q')$-th block of the Hessian of $f(\Ab)$ can be characterized as
\begin{align}
&\forall k, k' \in [l], \; q, q' \in [r] \nonumber\\
&\nabla_{\vecz{\Ab_{k'q'}}} \left(\nabla_{\vecz{\Ab_{kq}}} f(\Ab^*) \right) = \Qb_{k'q'}(\Ab^*)^\top\Qb_{kq}(\Ab^*), \label{eq: hessian_resnet}
\end{align}
where matrix $\Qb_{kq}$ is given in \cref{eq: Q_kq}.
We note that the above Hessian is evaluated at a global minimizer $\Ab^*$, and is hence different from the Hessian evaluated at the origin (i.e., $\Ab=0$) in \cite{Sihan_2016}.

Denote $\Qb(\Ab):= [\Qb_{11}(\Ab), \Qb_{12}(\Ab), \ldots, \Qb_{lr}(\Ab)]$. Then the entire Hessian matrix at any global minimizer can be characterized as
\begin{align}
\nabla_{\vecz{\Ab}}^2 f(\Ab^*) = \Qb(\Ab^*)^\top \Qb(\Ab^*).
\end{align}

The following theorem characterizes the regularity condition for the square loss function of linear residual networks.
\begin{thm}\label{thm: rc_resnet}
Consider $f(\Ab)$ of linear residual neural networks with $m=d$. Further consider a global minimizer $\Ab^*$, and let $\zeta = 2\max_{k\in[l]} \norm{\Wb_k^*}$ and $\tilde{\zeta} = 2\max_{k\in[l],q\in[r]} \|\Ab_{kq}^*\|$. Then, for any constant $\delta>0$, there exists a sufficiently small $\epsilon (\delta)$ such that any point $\Ab$ that satisfies $$\norm{\Qb(\Ab^*) \vecz{\Ab- \Ab^*}}_2 \ge \delta \norm{\vecz{\Ab- \Ab^*}}_2$$
	and within the neighborhood of $\Ab^*$ defined as $\{\Ab: \|\Ab_{kq}-\Ab_{kq}^*\|_F<\epsilon(\delta), \forall k\in[l], q\in[r] \}$, satisfies
	\begin{align}
	&\inner{\nabla_{\vecz{\Ab}} f(\Ab)}{\vecz{\Ab- \Ab^*}} \nonumber\\
	&\ge \alpha \norm{\nabla_{\vecz{\Ab}} f(\Ab)}_2^2 + \beta \norm{\vecz{\Ab- \Ab^*}}_2^2,
	\end{align}
	where $\alpha = \gamma / (lr\tilde{\zeta}^{2(r-1)} \zeta^{2(l-1)} \norm{\Xb}^2)$ and $\beta = (1-\gamma)\delta^2/2$ with any $\gamma \in (0,1)$.
\end{thm}

Similarly to the regularity condition for linear networks, the regularity condition for linear residual networks holds only along the directions of $\Ab$ such that $\frac{\norm{\Qb(\Ab^*) \vecz{\Ab- \Ab^*}}_2}{\norm{\vecz{\Ab- \Ab^*}}_2} \ge \delta$. However, the parametrization of $\Qb(\Ab^*)$ of linear residual networks is different from that of $\Gb(\Wb^*)$ of linear networks. To illustrate, consider a simplified setting of the linear residual network where the shortcut depth is $r=1$. Then, one has 
$$\Qb_{k}(\Ab^*) = \left(\Wb^*_{k-1}\ldots\Wb^*_1\Xb \right)^\top \otimes  \left(\Wb^*_l\ldots\Wb^*_{k+1} \right).$$ Although it takes the same form as $\Gb(\Wb^*)$ of the linear network, the reparameterization of $\Wb_k^* = \Ib + \Ab_k^*$ keeps $\eta_{\min}(\Wb_k^*)$ away from zero when $\norm{\Ab_k^*}$ is small. This enlarges $\eta_{\min}(\Qb_{k}(\Ab^*))$ so that the constraint $\frac{\norm{\Qb(\Ab^*) \vecz{\Ab- \Ab^*}}_2}{\norm{\vecz{\Ab- \Ab^*}}_2} \ge \delta$ can be satisfied along a wider range of directions. In this way, $\Ab^*$ attracts the optimization iteration path to converge along a wider range of directions in the neighborhood of the origin.


\subsection{Nonlinear Neural Networks with One Hidden Layer}
The Hessian of general nonlinear networks can take complicated forms, analyzing which is typically not trackable. Here, we consider nonlinear neural networks with one hidden layer and focus on a simplified setting where $\mathrm{range}(\sigma) = \RR$. In this case, $\sigma(\Wb_1\Xb)$ can realize any square matrix and hence the global minimal value of $g(\Wb)$ is zero. Consequently, the Hessian of $g(\Wb)$ at a global minimizer $\Wb^*$ takes the following form
\begin{align}
&\nabla_{\vecz{\Wb}}^2 g(\Wb^*) = \Hb(\Wb^*)^\top \Hb(\Wb^*), \label{eq: nonlinear_hessian}
\end{align}
where
\begin{align}
&\Hb^\top = 
\begin{bmatrix}
(\Xb \otimes \Ib)\sigma'(\mathrm{diag}(\vecz{\Wb_1^*\Xb})) (\Ib\otimes (\Wb_2^*)^\top)  \\
\sigma(\Wb_1^*\Xb) \otimes \Ib 
\end{bmatrix}.
\nonumber
\end{align}

The following theorem characterizes the regularity condition for the square loss function of nonlinear neural networks.
\begin{thm}\label{thm: rc_nonlinear}
Consider $g(\Wb)$ of one-hidden-layer nonlinear neural networks with $m=d$ and $\mathrm{range}(\sigma) = \RR$. Further consider a global minimizer $\Wb^*$ of $g(\Wb)$, and let $\zeta = 2\max \{\norm{\sigma(\Wb_1^* \Xb)}, \norm{\Wb_2^*}, \norm{\sigma'(\Wb_1^*\Xb)}_{\infty}\}$. Then there exists a sufficiently small $\epsilon (\delta)$ such that any point $\Wb$ that satisfies 
	$$\norm{\Hb(\Wb^*) \vecz{\Wb- \Wb^*}}_2 \ge \delta \norm{\vecz{\Wb- \Wb^*}}_2$$ 
	and within the neighborhood of $\Wb^*$ defined as $\{\Wb: \|\Wb_k-\Wb_k^*\|_F<\epsilon(\delta), \forall k\in[2] \}$ satisfies 
	\begin{align}
	&\langle\nabla_{\vecz{\Wb}} g(\Wb), \vecz{\Wb- \Wb^*} \rangle\nonumber\\
	&\ge \alpha \norm{\nabla_{\vecz{\Wb}} g(\Wb)}_2^2 + \beta \norm{\vecz{\Wb- \Wb^*}}_2^2, \label{eq: regular1}
	\end{align}
	where $\alpha = \gamma/\max\{\norm{\Xb}^2 \zeta^4, \zeta^2\}$ and $\beta = (1-\gamma) \delta^2/2$ for any $\gamma\in (0,1)$.
\end{thm}
Thus, nonlinear neural networks with one hidden layer also have an amenable landscape near the global minimizers that attracts gradient iterates to converge along the directions restricted by $\Hb(\Wb^*)$. 

\section{Conclusion}
In this paper, we explored two landscape properties for square losses of three types of neural networks: linear, linear residual, and one-hidden-layer nonlinear networks. We focused on the scenario with square parameter matrices, which allows us to characterize the explicit form of the global minimizers for these networks up to an equivalence class. Moreover, the characterization of global minimizers helps to establish the gradient dominance condition and the regularity condition for these networks in the neighborhood of their global minimizers under certain conditions. 
Along this direction, many interesting questions can be further asked, and are worth exploration in the future. For example, can we generalize the existing results for deep linear and shallow nonlinear networks to deep nonlinear networks? How can we further exploit the information about the higher-order derivatives of loss functions to understand the landscape of loss functions? Furthermore, it is interesting to exploit these gradient dominance condition and regularity condition in the convergence analysis of optimization algorithms applied to deep learning networks. The ultimate goal is to develop the theory that effectively exploits the properties of loss functions to guide the design of optimization algorithms for deep neural networks in practice.

{
	\bibliographystyle{apalike}
	\bibliography{./ref}
	\balance
}

\clearpage
\onecolumn
\appendix{
	{\centering\Large \textbf{Supplementary Materials}}

%
\section{Proof of Main Reults}

\section*{Proof of \Cref{thm: optima_linear}}
 We first consider the two-layer linear network $h(\Ab, \Bb):= \frac{1}{2} \norm{\Ab\Bb\Xb - \Yb}_F^2$, where $\Ab \in \RR^{d\times d}, \Bb\in \RR^{d\times d}$. Then the global minimizers are fully (sufficiently and necessarily) characterized by \cite[Fact 4]{Baldi_1989_1} as
\begin{align}
\Ab = \Ub \Cb,\qquad \Bb = \Cb^{-1}\Ub^\top \Sigmab_{\Xb\Yb}^\top \Sigmab_{\Xb\Xb}^{-1}, \label{eq: baldi}
\end{align}
where $\Cb$ is an arbitrary invertible matrix and $\Ub = [\ub_1,\ldots, \ub_d]$ is the matrix formed by the eigenvectors that correspond to the top $d$ eigenvalues of $\Sigmab = \Sigmab_{\Xb\Yb}^\top \Sigmab_{\Xb\Xb}^{-1} \Sigmab_{\Xb\Yb}$. Moreover, the global minimum value is given by $\min_{\Ab,\Bb} h(\Ab, \Bb) = \mathrm{Tr}(\Sigmab_{\Yb\Yb}) - \sum_{i=1}^{d} \lambda_i$, where $\lambda_i, i\in [d]$ are the top $d$ eigenvalues of $\Sigmab$.

Now consider the deep linear neural network with $l-1$ layers. Since all parameter matrices are square, $\mathrm{range}(\Wb_l\ldots\Wb_1) = \mathrm{range}(\Ab\Bb)$ and hence $\min_{\Ab,\Bb} h(\Ab, \Bb) = \min_{\Wb} h(\Wb)$. Thus, for any $k=2,\ldots,l$, we can treat $\Ab := \Wb_l\ldots\Wb_k$ and $\Bb:= \Wb_{k-1}\ldots\Wb_1$ and apply the characterization in \cref{eq: baldi} (due to its uniqueness). We then obtain that any $\Wb_l\ldots\Wb_k$, $\Wb_{k-1}\ldots\Wb_1$ that achieves the global optimum of $h(\Wb)$ is fully (sufficiently and necessarily) characterized by
\begin{align}
	\forall k=2,\ldots, l,\quad &\Wb_l\ldots\Wb_k = \Ub \Cb_{k},\label{eq: 1}\\
	&\Wb_{k-1}\ldots\Wb_1=\Cb_{k}^{-1}\Ub^\top \Sigmab_{\Xb\Yb}^\top \Sigmab_{\Xb\Xb}^{-1},\label{eq: 2}
\end{align}
where $\Cb_k$ are arbitrary invertible matrices. Note that \cref{eq: 1} with $k=l,l-1$ gives 
\begin{align}
	\Wb_l = \Ub \Cb_{l},\quad \Wb_l\Wb_{l-1} = \Ub \Cb_{l-1},
\end{align}
which further imply
\begin{align}
	\Ub\Cb_l\Wb_{l-1} = \Ub\Cb_{l-1}.
\end{align}
Multiplying both sides by $\Ub^\top$ and noting that $\Ub^\top\Ub = \Ib$, one can solve for $\Wb_{l-1}$ as $\Wb_{l-1} = \Cb_l^{-1} \Cb_{l-1}$. We then apply \cref{eq: 1} inductively and obtain that
\begin{align}
	\Wb_l = \Ub\Cb_l,\;\ldots,\; \Wb_k = \Cb_{k+1}^{-1} \Cb_k,\;\ldots,\; \Wb_1 = \Cb_2^{-1}\Ub^\top \Sigmab_{\Xb\Yb}^\top \Sigmab_{\Xb\Xb}^{-1}. \label{eq: 12}
\end{align}
One can verify that such a solution also satisfy the conditions in \cref{eq: 2} for all $k=2,\ldots,l$, and hence fully characterizes global minimizers. Clearly, the global minimizers characterized by \cref{eq: 12} belong to a unique equivalence class.


\section*{Proof of \Cref{thm: form_resnet}}
Consider any full-rank global minimizer $\Ab^*$. Since $m = d$, the global minimal value of $f(\Ab)$ is zero, and the fact that $\Ab^*$ minimizes $f(\Ab)$ implies that the corresponding $\Wb^*$ minimizes $h(\Wb)$. Thus, by the characterization in \Cref{thm: optima_linear}, we conclude that $\Wb^*$ is of full rank, and must be characterized as
\begin{align}
\Wb_l^* = \widehat{\Ub} \widehat{\Cb_l},\ldots, \Wb_{k}^* = \widehat{\Cb}_{k+1}^{-1}\widehat{\Cb_k},\ldots, \Wb_1^* = \widehat{\Cb_2}^{-1}\widehat{\Ub}^\top \Sigmab_{\Xb\Yb} \Sigmab_{\Xb\Xb}^{-1},
\end{align}
where $\widehat{\Cb_2},\ldots, \widehat{\Cb_l}$ are arbitrary invertible matrices and $\widehat{\Ub} = [\hat{\ub_1},\ldots, \hat{\ub_d}]$ is the matrix formed by all the eigenvectors of $\Sigmab$.

Since by definition $\Ab_{kr}^*\ldots\Ab_{k1}^* - (\Wb_k^* - \Ib) = 0$, $\Ab_{k1}^*,\ldots, \Ab_{kr}^*$ minimizes the loss $\frac{1}{2}\norm{\Ab_{kr}\ldots\Ab_{k1} - (\Wb_k^* - \Ib)}_F^2$ of a linear network. And $\Wb_k^*- \Ib$ is of full rank since all $\Ab_{kq}^*$ are of full rank. We want to apply  \Cref{thm: optima_linear} again to characterize the form of $\Ab_{kq}^*$ for $q\in [r]$. Note that $(\Wb_k^* - \Ib)(\Wb_k^* - \Ib)^\top$ may not have distinct eigenvalues. Hence by the remark in the proof of \cite[Fact 4]{Baldi_1989_1}, the matrix $\Ub$ formed by the eigenvectors should be generalized to $\Ub\Vb$, where $\Vb$ is block diagonal with each block being an orthogonal matrix (where the dimension is determined by the multiplicities of the eigenvalues). Then we can characterize $\Ab_{kq}^*$ as
\begin{align}
\Ab_{kr}^* = \Ub\Vb_r \Cb_r,\ldots,\; \Ab_{kq}^* = \Cb_{q+1}^{-1}\Vb_{q+1}^{-1} \Vb_q\Cb_q,\ldots,\; \Ab_{k1}^* = \Cb_2^{-1}\Vb_2^{-1}\Ub^\top (\Wb_k^* - \Ib),
\end{align}
where $\Cb_2,\ldots, \Cb_l$ are arbitrary invertible matrices, $\Vb_2,\ldots, \Vb_l$ are orthogonal block diagonal matrices, and $\Ub =  [\ub_1,\ldots, \ub_d]$ is the matrix formed by all the eigenvectors of $(\Wb_k^* - \Ib)(\Wb_k^* - \Ib)^\top$.
It can be observed that combining $\Vb_q\Cb_q$ into one invertible transformation $\Cb_q$ does not change the generalization of the form, which thus yields the desired result.

In the case of $r=1$, the index $q$ is absent and $\Wb_k^* = \Ib + \Ab_k^*$ for all $k\in[l]$. Consider any global minimizer $\Ab^*$ (not necessarily of full rank). Since the global minimum of $f(\Ab)$ is zero, the fact that $\Ab^*$ minimizes $f(\Ab)$ implies that the corresponding $\Wb^*$ minimizes $h(\Wb)$, and such $\Wb^*$ is characterized by \cref{eq:wmatrix}. We thus obtain $\Ab_k^* = \Wb_k^* - \Ib$.

\section*{Proof of \Cref{prop: globmin_nonlinear}} 
Consider the loss function $\tfrac{1}{2} \norm{\Wb_2 \sigma(\Wb_{1} \Xb) - \Yb}_F^2$ of a nonlinear network and the loss function $\tfrac{1}{2} \norm{\widetilde{\Wb}_2 \widetilde{\Wb}_{1} \Xb - \Yb}_F^2$ of the corresponding linear network. Let $(\Wb_2^*, \Wb_1^*)$ be a global minimizer of the nonlinear network (achieves zero loss since $m = d$). Then one can verify that $(\widetilde{\Wb}_2^*, \widetilde{\Wb}_1^*)$ achieves zero loss of the linear network with $\widetilde{\Wb}_2^* = \Wb_2^*, \widetilde{\Wb}_1^* = \sigma(\Wb_{1}^* \Xb)\Xb^{-1}$. Such $(\widetilde{\Wb}_2^*, \widetilde{\Wb}_1^*)$ is a global minimizer of the linear network, and can be fully characterized by \Cref{thm: optima_linear}. Moreover, since $\mathrm{range}(\sigma) = \RR$, we can take the inverse of $\sigma$ and conclude that $\Wb_1^* = \sigma^{-1}(\widetilde{\Wb}_1^* \Xb)\Xb^{-1}$. 	We note that the inverse function should be understood as $\sigma^{-1}(y) := \{x~|~\sigma(x) = y \}$, and is well defined since $\mathrm{range}(\sigma) = \RR$.

\section*{Proof of \Cref{thm: gd_condition}}
We first consider any full-rank point $\Wb$, i.e., $\eta_{\min}(\Wb_k)>0$ for all $k\in [l]$. We obtain for all $k\in[l]$ that
	\begin{align}
	\norm{\nabla_{\vecz{\Wb_k}} h(\Wb)}_2  &= \norm{\Gb_k(\Wb)^\top \vecz{\eb}}_2 \\
	&\ge \eta_{\min}(\Gb_k(\Wb)) \norm{\vecz{\eb}}_2 \\
	&= \eta_{\min}\left((\Wb_{k-1}\ldots\Wb_1\Xb)^\top\otimes (\Wb_l\ldots\Wb_{k+1})\right) \norm{\vecz{\eb}}_2\\
	&\overset{(i)}{=} \eta_{\min}(\Wb_{k-1}\ldots\Wb_1\Xb) \eta_{\min}(\Wb_l\ldots\Wb_{k+1}) \norm{\vecz{\eb}}_2,
	\end{align}
	where (i) follows from the fact that $\eta_{\min}(\Ab\otimes \Bb) = \eta_{\min}(\Ab)\eta_{\min}(\Bb)$. Thus, by summing over $k\in[l]$ we obtain
	\begin{align}
		\sum_{k=1}^{l} \norm{\nabla_{\vecz{\Wb_k}} h(\Wb)}_2^2 &\ge \sum_{k=1}^{l} \eta_{\min}^2(\Wb_{k-1}\ldots\Wb_1\Xb) \eta_{\min}^2(\Wb_l\ldots\Wb_{k+1})\norm{\vecz{\eb}}_2^2 \\
		&= 2h(\Wb) \left(\sum_{k=1}^{l} \eta_{\min}^2(\Wb_{k-1}\ldots\Wb_1\Xb) \eta_{\min}^2(\Wb_l\ldots\Wb_{k+1})\right).  
	\end{align}
	After rearranging, we obtain that
	\begin{align}
		h(\Wb) \le \lambda_h \norm{\nabla_{\vecz{\Wb}} h(\Wb)}_2^2, \label{eq: 10}
	\end{align}
	where $\lambda_h = 1/\left(2\sum_{k=1}^{l} \eta_{\min}^2(\Wb_{k-1}\ldots\Wb_1\Xb) \eta_{\min}^2(\Wb_l\ldots\Wb_{k+1}) \right)$.
	
	Now consider any global minimizer $\Wb^*$ of $h(\Wb)$. \Cref{thm: optima_linear} guarantees that $\Wb^*$ is of full rank under the assumption of the theorem. Consider any point $\Wb$ in the neighborhood of $\Wb^*$ defined as $\{\Wb:\|\Wb_k-\Wb_k^*\|<\tau,~ \forall k\in [l]\}$. Then, by Weyl's inequality \cite[page 178, Theorem 3.3.16]{Horn_1986}, we have for all $k\in [l]$
	\begin{align}
		\eta_{\min}(\Wb_k) = \eta_{\min}(\Wb_k^* + \Wb_k - \Wb_k^*) \ge \eta_{\min}(\Wb_k^*) - \norm{\Wb_k - \Wb_k^*} \ge 2\tau - \tau = \tau >0,
	\end{align}
	where $\tau$ is defined in the statement of the theorem.
	 Thus, $\Wb$ is of full rank and hence satisfies \cref{eq: 10}, where $\lambda_h$ can be further upper bounded as
	 \begin{align}
	 	\lambda_h &= \left(2\sum_{k=1}^{l} \eta_{\min}^2(\Wb_{k-1}\ldots\Wb_1\Xb) \eta_{\min}^2(\Wb_l\ldots\Wb_{k+1}) \right)^{-1} \\
	 	&\le (2l\tau^{2(l-1)}\eta_{\min}^2(\Xb))^{-1},
	 \end{align}
where we use the fact that $\eta_{\min}(\Ab\Bb) \ge \eta_{\min}(\Ab) \eta_{\min}(\Bb)$. Lastly, observe that $h(\Wb^*) = 0$ since $m = d$.

\section*{Proof of \Cref{thm: gd_resnet_1}}

The proof follows the argument similar to that for \Cref{thm: gd_condition}. We first consider a full-rank point $\Ab$ such that its corresponding $\Wb$ is also of full rank. Then for all $k\in [l], q\in [r]$
\begin{align}
\norm{\nabla_{\vecz{\Ab_{kq}}} f(\Ab)}_2 &= \norm{\Qb_{kq}(\Ab)^\top \vecz{\eb}}_2 \\
&\ge \eta_{\min}(\Qb_{kq}(\Ab))\vecz{\eb}_2 \\
&\overset{(i)}{\ge} \eta_{\min}\left(\Ab_{k(q-1)}\ldots\Ab_{k1}\right)  \eta_{\min}\left(\Ab_{kr}\ldots\Ab_{k(q+1)}\right) \\
&\qquad \cdot\eta_{\min}\left(\Wb_{k-1}\ldots\Wb_1\Xb \right)  \eta_{\min}\left(\Wb_l\ldots\Wb_{k+1} \right) \norm{\vecz{\eb}}_2, 
\end{align}
where (i) follows from the form of $\Qb_{kq}(\Ab)$.
Note that $f(\Ab) = \frac{1}{2}\norm{\eb(\Ab)}_2^2$. Then an argument similar to that of \Cref{thm: gd_condition} yields 
\begin{align}
f(\Ab) \le \lambda_f \norm{\nabla_{\vecz{\Ab}} f(\Ab)}_2^2, \label{eq: 11}
\end{align}
where
\begin{align}
\lambda_{f} &=  \Bigg(2\sum_{k,q}\eta_{\min}^2\left(\Ab_{k(q-1)}\ldots\Ab_{k1}\right)  \eta_{\min}^2\left(\Ab_{kr}\ldots\Ab_{k(q+1)}\right) \\
&\qquad\qquad\quad \cdot\eta_{\min}^2\left(\Wb_{k-1}\ldots\Wb_1 \Xb \right)  \eta_{\min}^2\left(\Wb_l\ldots \Wb_{k+1} \right)\Bigg)^{-1}.
\end{align}

Now consider a global minimizer $\Ab^*$ of $f(\Ab)$. Since $f(\Ab^*)=0$, the fact that $\Ab^*$ minimizes $f(\Ab)$ implies that the corresponding $\Wb^*$ minimizes $h(\Wb)$. Thus, \Cref{thm: optima_linear} guarantees that $\Wb^*$ is also of full rank. Consider any point in the neighborhood of $\Ab^*$ defined as $\{\Ab: \|\Ab_{kq}-\Ab_{kq}^*\|<\min\{\hat{\tau}, \tilde{\tau}\}, \forall k\in [l], q\in [r]\}$, and we have 
\begin{align}
\eta_{\min}(\Ab_{kq}) \ge \eta_{\min}(\Ab_{kq}^*) - \norm{\Ab_{kq} - \Ab_{kq}^*} \ge 2\tilde{\tau} - \tilde{\tau} = \tilde{\tau}>0.
\end{align}
Since such a point is also contained in the neighborhood $\{\Ab: \|\Ab_{kq}-\Ab_{kq}^*\|<\hat{\tau}, \forall k\in [l], q\in [r]\}$, the definition of $\hat{\tau}$ implies that for all $k\in[l]$
\begin{align}
\eta_{\min}(\Wb_{k}) \ge \eta_{\min}(\Wb_{k}^*) - \norm{\Wb_{k} - \Wb_{k}^*} \ge 2\tau - \tau = \tau>0.
\end{align}  
Thus, $\Ab$ and the corresponding $\Wb$ are of full rank and hence satisfies \cref{eq: 11}, where the parameter $\lambda_f$ can be further upper bounded as 
\begin{align}
\lambda_{f} &=  \Bigg(2\sum_{k,q}\eta_{\min}^2\left(\Ab_{k(q-1)}\ldots\Ab_{k1}\right)  \eta_{\min}^2\left(\Ab_{kr}\ldots\Ab_{k(q+1)}\right) \\
&\qquad\qquad\quad\cdot\eta_{\min}^2\left(\Wb_{k-1}\ldots\Wb_1 \Xb \right)  \eta_{\min}^2\left(\Wb_l\ldots \Wb_{k+1} \right)\Bigg)^{-1} \\
&\le \left(2lr\tilde{\tau}^{2(r-1)} \tau^{2(l-1)} \eta_{\min}^2(\Xb) \right)^{-1},
\end{align}
where we use the fact that $\eta_{\min}(\Ab\Bb) \ge \eta_{\min}(\Ab)\eta_{\min}(\Bb)$.

\section*{Proof of \Cref{thm: gd_nonlinear}}
Applying the derivatives in \cref{eq: nonlinear_grad_1,eq: nonlinear_grad_2} (the detailed calculations are provided in \Cref{app:derivative}), we obtain that for any $\Wb$
\begin{align*}
\norm{\nabla_{\vecz{\Wb}} g(\Wb)}_2^2 &\ge \norm{\nabla_{\vecz{\Wb_2}} g(\Wb)}_2^2 \\
&= \norm{(\sigma(\Wb_1\Xb) \otimes \Ib) \vecz{\eb}}_2^2 \\
&\ge \eta_{\min}^2(\sigma(\Wb_1\Xb) \otimes \Ib) \norm{\vecz{\eb}}_2^2 \\
&= 2\eta_{\min}^2(\sigma(\Wb_1\Xb)) g(\Wb).
\end{align*}
Consider any $\{\Wb: \norm{\sigma(\Wb_1\Xb) - \sigma(\Wb_1^*\Xb)}\le \tau \}$, by Weyl's inequality we obtain that
\begin{align*}
\eta_{\min}(\sigma(\Wb_1\Xb)) \ge \eta_{\min}(\sigma(\Wb_1^*\Xb)) - \norm{\sigma(\Wb_1\Xb) - \sigma(\Wb_1^*\Xb)} \ge 2\tau - \tau = \tau >0.
\end{align*}
Thus, we conclude that 
\begin{align*}
g(\Wb) \le \lambda_g \norm{\nabla_{\vecz{\Wb}} g(\Wb)}_2^2,
\end{align*}
where $\lambda_g = (2\eta_{\min}^2(\sigma(\Wb_1\Xb)))^{-1} \le (2\tau^2)^{-1}$. Lastly, observe that $g(\Wb^*) = 0$ since $m = d$ and $\mathrm{range}(\sigma) = \RR$.

\section*{Proof of \Cref{thm: rc_condition}}

Consider any $\Wb$ and any global minimizer $\Wb^*$. The Peano form of the Taylor expansion of $h(\Wb)$ at $\Wb^*$ gives
	\begin{align}
		h(\Wb) &= h(\Wb^*) + \vecz{\Wb- \Wb^*}^\top \nabla_{\vecz{\Wb}} h(\Wb^*)\nonumber\\
		&\quad+ \tfrac{1}{2} \vecz{\Wb- \Wb^*}^\top \nabla_{\vecz{\Wb}}^2 h(\Wb^*) \vecz{\Wb- \Wb^*} + o(\norm{\vecz{\Wb- \Wb^*}}_2^2).
	\end{align}
	Here $\frac{o(\norm{\wb}_2^2)}{\norm{\wb}_2^2}\to 0$ as $\norm{\wb}_2\to 0$.
	Note that $h(\Wb^*) = 0$, $\nabla_{\vecz{\Wb}} h(\Wb^*) = \bm{0}$ and $\nabla_{\vecz{\Wb}}^2 h(\Wb^*) = \Gb(\Wb^*)^\top \Gb(\Wb^*).$ We then obtain that
	\begin{align}
	h(\Wb) = \tfrac{1}{2} \norm{\Gb(\Wb^*) \vecz{\Wb- \Wb^*}}_2^2 + o(\norm{\vecz{\Wb- \Wb^*}}_2^2).\label{eq: 3}
	\end{align}
	Now consider the function $\ell(\Wb):= \inner{\nabla_{\vecz{\Wb}} h(\Wb)}{\vecz{\Wb- \Wb^*}}$. We observe that
	\begin{align}
		\nabla_{\vecz{\Wb}} \ell(\Wb^*)&= \nabla_{\vecz{\Wb}} [(\nabla_{\vecz{\Wb}} h(\Wb))^\top\vecz{\Wb- \Wb^*}] \bigg|_{\Wb = \Wb^*}\\
		&= \left[\left(\nabla_{\vecz{\Wb}}^2 h(\Wb)\right) \vecz{\Wb- \Wb^*}+ \nabla_{\vecz{\Wb}} h(\Wb)\right]\bigg|_{\Wb = \Wb^*}\\
		&=0.
	\end{align}
	Furthermore,
	\begin{align}
		\nabla_{\vecz{\Wb}}^2 \ell(\Wb^*) &= \nabla_{\vecz{\Wb}}\left[\left(\nabla_{\vecz{\Wb}}^2 h(\Wb)\right) \vecz{\Wb- \Wb^*}+ \nabla_{\vecz{\Wb}} h(\Wb)\right]\bigg|_{\Wb = \Wb^*}   \\
		&= \left[\left(\nabla_{\vecz{\Wb}} \nabla_{\vecz{\Wb}}^2 h(\Wb)\right) \vecz{\Wb- \Wb^*} + 2\nabla_{\vecz{\Wb}}^2 h(\Wb) \right]\bigg|_{\Wb = \Wb^*} \\
		&= 2\nabla_{\vecz{\Wb}}^2 h(\Wb^*).
	\end{align}
	Hence, $\ell(\Wb)$ has the same Hessian (up to a scaling factor) as $h(\Wb)$ at the global minimizer $\Wb^*$. Then,
	its Peano form of Taylor expansion is given by
	\begin{align}
		\ell(\Wb) = \norm{\Gb(\Wb^*) \vecz{\Wb- \Wb^*}}_2^2 + o(\norm{\vecz{\Wb- \Wb^*}}_2^2). \label{eq: 4}
	\end{align}
	Consider a sufficiently small $\epsilon(\delta)$-neighborhood of $\Wb^*$, we have $\norm{\Wb_k} \le \norm{\Wb_k - \Wb_k^*} + \norm{\Wb_k^*} \le \frac{\zeta}{2} + \frac{\zeta}{2} = \zeta$ for all $k\in [l]$. 
	
	On the other hand, by \cref{eq:gradient_linear}, we obtain that (by summing over $k\in [l]$)
	\begin{align}
	\sum_{k=1}^{l}\norm{\nabla_{\vecz{\Wb_k}} h(\Wb)}_2^2  &= \sum_{k=1}^{l}\norm{\Gb_k(\Wb)^\top \vecz{\eb}}_2^2 \\
	&\le \sum_{k=1}^{l}\norm{\Gb_k(\Wb)}^2 \norm{\vecz{\eb}}_2^2 \\
	&= \sum_{k=1}^{l}\norm{(\Wb_{k-1}\ldots\Wb_1\Xb)^\top\otimes (\Wb_l\ldots\Wb_{k+1})}^2 \norm{\vecz{\eb}}_2^2\\
	&\overset{(i)}{=} \sum_{k=1}^{l}\norm{\Wb_{k-1}\ldots\Wb_1\Xb}^2 \norm{\Wb_l\ldots\Wb_{k+1}}^2 \norm{\vecz{\eb}}_2^2  \\
	&\le l\zeta^{2(l-1)}\norm{\Xb}^2 \norm{\vecz{\eb}}_2^2, \label{eq: 5}
	\end{align}
	where (i) follows from the fact that $\norm{\Ab\otimes \Bb} = \norm{\Ab}\norm{\Bb}$.
	Note that by assumption, $\norm{\Gb(\Wb^*) \vecz{\Wb- \Wb^*}}_2^2 \ge \delta^2 \norm{\vecz{\Wb- \Wb^*}}_2^2$. Then comparing \cref{eq: 3} and \cref{eq: 4} yields that, for any $\gamma \in (0,1)$
	\begin{align}
		\ell(\Wb) &= (\gamma + 1 - \gamma) \norm{\Gb(\Wb^*) \vecz{\Wb- \Wb^*}}_2^2 + o(\norm{\vecz{\Wb- \Wb^*}}_2^2) \\
		&= 2\gamma h(\Wb)  + (1 - \gamma) \norm{\Gb(\Wb^*) \vecz{\Wb- \Wb^*}}_2^2 + o(\norm{\vecz{\Wb- \Wb^*}}_2^2) \\
		&= \gamma \norm{\eb(\Wb)}_2^2  + (1 - \gamma) \norm{\Gb(\Wb^*) \vecz{\Wb- \Wb^*}}_2^2 + o(\norm{\vecz{\Wb- \Wb^*}}_2^2) \\
		&\overset{(i)}{\ge} \alpha \norm{\nabla_{\vecz{\Wb}} h(\Wb)}_2^2 + \beta \norm{\vecz{\Wb- \Wb^*}}_2^2 + o(\norm{\vecz{\Wb- \Wb^*}}_2^2),
	\end{align}
	where (i) follows from \cref{eq: 5} with $\alpha = \gamma/(l\zeta^{2(l-1)} \norm{\Xb}^2)$ and $\beta = (1-\gamma) \delta^2.$ Finally, by considering a sufficiently small $\epsilon(\delta)$-neighborhood of $\Wb^*$, the higher order term $o(\norm{\vecz{\Wb- \Wb^*}}_2^2)$ can be removed by letting $\beta = (1-\gamma)\delta^2/2$.

\section*{Proof of \Cref{thm: rc_resnet}}
The proof follows the argument similar to that for \Cref{thm: rc_condition}, and we present an outline of the proof.

First, note that $\nabla_{\vecz{\Ab}} f(\Ab^*)=0$ and $\nabla_{\vecz{\Ab}}^2 f(\Ab^*) = \Qb(\Ab^*)^\top\Qb(\Ab^*)$. Then the Peano form of the Taylor expansion of $f(\Ab)$ at $\Ab^*$ is given by
\begin{align}
f(\Ab) = \tfrac{1}{2} \norm{\Qb(\Ab^*) \vecz{\Ab- \Ab^*}}_2^2 + o(\norm{\vecz{\Ab- \Ab^*}}_2^2).\label{eq: 7}
\end{align}
Now consider the function $\ell(\Ab):= \inner{\nabla_{\vecz{\Ab}} f(\Ab)}{\vecz{\Ab- \Ab^*}}$. Similarly to the proof of \Cref{thm: rc_condition}, one can show that $\nabla_{\vecz{\Ab}} \ell(\Ab^*) = 0$, $\nabla_{\vecz{\Ab}}^2 \ell(\Ab^*) = 2\nabla_{\vecz{\Ab}}^2 f(\Ab^*)$, and hence the Peano form of the Taylor expansion of $\ell(\Ab)$ is given by
\begin{align}
\ell(\Ab) = \norm{\Qb(\Ab^*) \vecz{\Ab- \Ab^*}}_2^2 + o(\norm{\vecz{\Ab- \Ab^*}}_2^2). \label{eq: 8}
\end{align}
Consider a sufficiently small $\epsilon(\delta)$ neighborhood of $\Ab^*$, which also implies a sufficiently small neighborhood of $\Wb^*$ (since $\norm{\Wb-\Wb^*}\to 0$ as $\norm{\Ab-\Ab^*}\to 0$). We then have $\norm{\Ab_{kq}} \le \norm{\Ab_{kq} - \Ab_{kq}^*} + \norm{\Ab_{kq^*}} \le \frac{\tilde{\zeta}}{2} +\frac{\tilde{\zeta}}{2} = \tilde{\zeta}$ and similarly $\norm{\Wb_k} \le \zeta$. On the other hand, by \cref{eq: 6}, we obtain that
\begin{align}
\sum_{kq}\norm{\nabla_{\vecz{\Ab_{kq}}} f(\Ab)}_2^2 &\le \sum_{kq}\norm{\Qb_{kq}(\Ab)}^2 \norm{\eb}_2^2 \\
&\le \sum_{kq} \norm{\Ab_{k(q-1)}\ldots\Ab_{k1}}^2  \norm{\Ab_{kr}\ldots\Ab_{k(q+1)}}^2 \\
&\qquad \cdot\norm{\Wb_{k-1}\ldots\Wb_1\Xb}^2  \norm{\Wb_l\ldots\Wb_{k+1}}^2 \norm{\vecz{\eb}}_2^2 \\
&\le lr\tilde{\zeta}^{2(r-1)}\zeta^{2(l-1)}\norm{\Xb}^2 \norm{\vecz{\eb}}_2^2. \label{eq: 9}
\end{align}
Note that by assumption, $\norm{\Qb(\Ab^*) \vecz{\Ab- \Ab^*}}_2^2 \ge \delta^2 \norm{\vecz{\Ab- \Ab^*}}_2^2$. 
Then comparison of \cref{eq: 7,eq: 8} yields that, for any $\gamma \in (0,1)$,
\begin{align}
\ell(\Ab) &= (\gamma + 1 - \gamma) \norm{\Qb(\Ab^*) \vecz{\Ab- \Ab^*}}_2^2 + o(\norm{\vecz{\Ab- \Ab^*}}_2^2) \\
&= 2\gamma f(\Ab)  + (1 - \gamma) \norm{\Qb(\Ab^*) \vecz{\Ab- \Ab^*}}_2^2 + o(\norm{\vecz{\Ab- \Ab^*}}_2^2) \\
&= \gamma \norm{\eb(\Ab)}_2^2  + (1 - \gamma) \norm{\Qb(\Ab^*) \vecz{\Ab- \Ab^*}}_2^2 + o(\norm{\vecz{\Ab- \Ab^*}}_2^2) \\
&\overset{(i)}{\ge} \alpha \norm{\nabla_{\vecz{\Ab}} f(\Ab)}_2^2 + \beta \norm{\vecz{\Ab- \Ab^*}}_2^2 + o(\norm{\vecz{\Ab- \Ab^*}}_2^2),
\end{align}
where (i) follows from \cref{eq: 9} with $\alpha = \gamma / (lr\tilde{\zeta}^{2(r-1)} \zeta^{2(l-1)} \norm{\Xb}^2)$ and $\beta = (1-\gamma)\delta^2$. Finally, by considering a sufficiently small $\epsilon(\delta)$ neighborhood of $\Ab^*$, the higher order term $o(\norm{\vecz{\Ab- \Ab^*}}_2^2)$ can be removed by letting $\beta = (1-\gamma)\delta^2/2$.

\section*{Proof of \Cref{thm: rc_nonlinear}}
 Consider a global minimizer $\Wb^*$ of $g(\Wb)$ which satisfies $g(\Wb^*) = 0$ and $ \nabla_{\vecz{\Wb}} g(\Wb^*) = \zero$.  Define the function $\ell(\Wb):= \inner{\nabla_{\vecz{\Wb}} g(\Wb)}{\vecz{\Wb- \Wb^*}}$. Then, following the similar argument of the proof of \Cref{thm: rc_condition}, we conclude that
 \begin{align*}
 	g(\Wb) &= \tfrac{1}{2} \norm{\Hb(\Wb^*) \vecz{\Wb- \Wb^*}}_2^2 + o(\norm{\vecz{\Wb- \Wb^*}}_2^2), \\
 	\ell(\Wb) &= \norm{\Hb(\Wb^*) \vecz{\Wb- \Wb^*}}_2^2 + o(\norm{\vecz{\Wb- \Wb^*}}_2^2). 
 \end{align*}
Consider a sufficiently small $\epsilon(\delta)$-neighborhood of $\Wb^*$, we have $\norm{\Wb_2} \le \norm{\Wb_2 - \Wb_2^*} + \norm{\Wb_2^*} \le \frac{\zeta}{2} + \frac{\zeta}{2} = \zeta$. Moreover, by continuity we have $\norm{\sigma(\Wb_1\Xb) } \le \norm{ \sigma(\Wb_1^*\Xb)} + \norm{\sigma(\Wb_1\Xb) - \sigma(\Wb_1^*\Xb)} \le \zeta$ and $\norm{\sigma'(\Wb_1\Xb) }_{\infty} \le \norm{ \sigma'(\Wb_1^*\Xb)}_{\infty} + \norm{\sigma'(\Wb_1\Xb) - \sigma'(\Wb_1^*\Xb)}_{\infty} \le \zeta$.  

On the other hand, \cref{eq: nonlinear_grad_1,eq: nonlinear_grad_2} imply that
\begin{align*}
	\norm{\nabla_{\vecz{\Wb_2}} g(\Wb)}_2^2 &\le \norm{\sigma(\Wb_1\Xb)}^2 \norm{\eb}_F^2 \le \zeta^2 \norm{\eb}_F^2
\end{align*}
and 
\begin{align*}
	\norm{\nabla_{\vecz{\Wb_1}} g(\Wb)}_2^2 &= \norm{\Xb}^2 \norm{\sigma'(\Wb_1\Xb) \circ (\Wb_2^\top \eb)}_F^2 \\
	&\le \norm{\Xb}^2 \norm{\sigma'(\Wb_1\Xb)}_{\infty}^2 \norm{\Wb_2^\top \eb}_F^2 \\
	&\le \norm{\Xb}^2 \norm{\sigma'(\Wb_1\Xb)}_{\infty}^2 \norm{\Wb_2}^2 \norm{\eb}_F^2 \\
	&\le \norm{\Xb}^2 \zeta^4 \norm{\eb}_F^2.
\end{align*}
Combining the above two inequalities, we further obtain that $\norm{\eb}_F^2 \ge \max\{\norm{\Xb}^2 \zeta^4, \zeta^2\}^{-1} \norm{\nabla_{\vecz{\Wb}} g(\Wb)}_2^2$.

Note that by assumption, $\norm{\Hb(\Wb^*) \vecz{\Wb- \Wb^*}}_2^2 \ge \delta^2 \norm{\vecz{\Wb- \Wb^*}}_2^2$. Then comparing $g(\Wb)$ and $\ell(\Wb)$ yields that, for any $\gamma \in (0,1)$
\begin{align}
\ell(\Wb) &= (\gamma + 1 - \gamma) \norm{\Gb(\Wb^*) \vecz{\Wb- \Wb^*}}_2^2 + o(\norm{\vecz{\Wb- \Wb^*}}_2^2) \\
&= 2\gamma h(\Wb)  + (1 - \gamma) \norm{\Gb(\Wb^*) \vecz{\Wb- \Wb^*}}_2^2 + o(\norm{\vecz{\Wb- \Wb^*}}_2^2) \\
&= \gamma \norm{\eb(\Wb)}_2^2  + (1 - \gamma) \norm{\Gb(\Wb^*) \vecz{\Wb- \Wb^*}}_2^2 + o(\norm{\vecz{\Wb- \Wb^*}}_2^2) \\
&\ge \alpha \norm{\nabla_{\vecz{\Wb}} g(\Wb)}_2^2 + \beta \norm{\vecz{\Wb- \Wb^*}}_2^2 + o(\norm{\vecz{\Wb- \Wb^*}}_2^2),
\end{align}
where $\alpha = \gamma/\max\{\norm{\Xb}^2 \zeta^4, \zeta^2\}$ and $\beta = (1-\gamma) \delta^2.$ Finally, by considering a sufficiently small $\epsilon(\delta)$-neighborhood of $\Wb^*$, the higher order term $o(\norm{\vecz{\Wb- \Wb^*}}_2^2)$ can be removed by letting $\beta = (1-\gamma)\delta^2/2$.

\section{Proof of the First-Order and Second-Order Derivatives}\label{app:derivative}	
	
\section*{Proof of First and Second Order Derivatives of $h(\Wb)$ in \cref{eq: grad_linear,eq: hessian_linear}}
We adopt the denominator layout convention for expressing derivatives. In specific, for a continuously differentiable function $f:\RR^d \to \RR$, its first-order derivative $\nabla_{\xb} f(\xb) \in \RR^{d\times 1}$ is defined as
\begin{align}
\nabla_{\xb} f(\xb) := \left[\tfrac{\partial f(\xb)}{\partial x_1}\quad  \tfrac{\partial f(\xb)}{\partial x_2}\quad \cdots\quad \tfrac{\partial f(\xb)}{\partial x_d}\right]^\top.
\end{align}
By this convention, the derivative is a column vector. Moreover, we frequently use the following composition rule, product rule and chain rule by denominator layout. Here, $\ub(\xb)$ and $ \vb(\xb)$ are vector valued functions with the same dimension and $\Ab$ is any dimension compatible matrix.
\begin{align}
\nabla_{\xb} \ub(\vb) =  (\nabla_{\xb} \vb)(\nabla_{\vb} \ub),\quad   \nabla_{\xb} \ub^\top \vb = (\nabla_{\xb}\ub)\vb + (\nabla_{\xb}\vb)\ub, \quad \nabla_{\xb} \Ab\xb = \Ab^\top.
\end{align}

Note that the loss function $h(\Wb)$ of linear network can be expressed as 
\begin{equation}
h(\Wb) = \tfrac{1}{2} \vecz{\eb(\Wb)}^\top \vecz{\eb(\Wb)}.
\end{equation}
Consider the following first-order derivative 
\begin{equation}
\nabla_{\vecz{\Wb}} h(\Wb) := \left[\left(\nabla_{\vecz{\Wb_1}} h(\Wb)\right)^\top \ldots \left(\nabla_{\vecz{\Wb_l}} h(\Wb)\right)^\top\right]^\top.
\end{equation} 
For any $k\in [l]$, we have
\begin{align}
\nabla_{\vecz{\Wb_k}} h(\Wb) &= \nabla_{\vecz{\Wb_k}} \tfrac{1}{2} \vecz{\eb}^\top \vecz{\eb} \\
&\overset{(i)}{=}\left(\nabla_{\vecz{\Wb_k}} \vecz{\eb} \right) \vecz{\eb} \\
&= \left[\nabla_{\vecz{\Wb_k}} \vecz{\Wb_l\Wb_{l-1}\ldots\Wb_1\Xb}  \right]\vecz{\eb} \\
&\overset{(ii)}{=} \left[\nabla_{\vecz{\Wb_k}} \left((\Wb_{k-1}\ldots\Wb_1\Xb)^\top \otimes (\Wb_l\ldots\Wb_{k+1})\right) \vecz{\Wb_k}  \right]\vecz{\eb} \\
&\overset{(iii)}{=} \left((\Wb_{k-1}\ldots\Wb_1\Xb)\otimes (\Wb_l\ldots\Wb_{k+1})^\top \right) \vecz{\eb} \\
&\overset{(iv)}{=}\Gb_k(\Wb)^{\top}\vecz{\eb}, \label{eq:gradient_linear}
\end{align}
where (i) follows from the product rule, (ii) follows because $\vecz{\Ab\Xb\Bb} = (\Bb^\top \otimes \Ab) \vecz{\Xb}$, (iii) follows from the chain rule and the fact that $(\Ab \otimes \Bb)^\top = (\Ab^\top \otimes \Bb^\top)$, and (iv) follows from the definition of $\Gb_k(\Wb): = (\Wb_{k-1}\ldots\Wb_1\Xb)^\top\otimes (\Wb_l\ldots\Wb_{k+1})$.

We next derive the Hessian of $h(\Wb)$ at a global minimizer $\Wb^*$. In the setting $d=m$ where all matrices are square, the global minimum of $h(\Wb)$ equals zero, i.e., $\eb(\Wb^*) = \bm{0}$. Then the $(k, k')$-block of the Hessian can be characterized as
\begin{align}
\nabla_{\vecz{\Wb_{k'}}} \left(\nabla_{\vecz{\Wb_k}} h(\Wb^*) \right) &= \nabla_{\vecz{\Wb_{k'}}} \left(\Gb_k^\top(\Wb^*) \vecz{\eb} \right) \\
&=  \left(\nabla_{\vecz{\Wb_{k'}}} \Gb_k(\Wb^*) \right)\vecz{\eb}  + \left(\nabla_{\vecz{\Wb_{k'}}} \vecz{\eb} \right)\Gb_k(\Wb^*) \\
&\overset{(i)}{=} 0 + \Gb_{k'}^\top(\Wb^*)\Gb_k(\Wb^*),
\end{align}
where (i) follows because $\eb(\Wb^*) = \bm{0}$ and uses the characterization of the first-order derivative in \cref{eq:gradient_linear}.

\section*{Proof of First and Second Order Derivatives of $f(\Ab)$ in \cref{eq: grad_resnet,eq: hessian_resnet}}
Note that $\Wb_k := \Ib + \Ab_{kr}\ldots\Ab_{k1}$.
For the $(k, q)$-th block of the gradient, we have
\begin{align}
\nabla_{\vecz{\Ab_{kq}}} f(\Ab) &= \nabla_{\vecz{\Ab_{kq}}} \tfrac{1}{2} \vecz{\eb}^\top \vecz{\eb} \\
&\overset{(i)}{=}\left(\nabla_{\vecz{\Ab_{kq}}}\vecz{\eb} \right) \vecz{\eb} \\
&\overset{(ii)}{=}  \left(\nabla_{\vecz{\Ab_{kq}}}\vecz{\Wb_k} \right) \left(\nabla_{\vecz{\Wb_{k}}}\vecz{\eb} \right) \vecz{\eb} \\
&\overset{(iii)}{=} \left[\left(\prod_{j=q\!-\!1}^1 \Ab_{kj}  \right)\!\!\otimes\!  \left(\prod_{j=r}^{q+1} \Ab_{kj}  \right)^\top  \right] \Gb_k(\Wb)^\top\vecz{\eb} \\
&\overset{(iv)}{=} \Qb_{kq}(\Ab)^\top \vecz{\eb}, \label{eq: 6}
\end{align}
where (i) follows from the product rule, (ii) follows from the composition rule, (iii) uses the result in \cref{eq:gradient_linear}, and (iv) is due to the definition of $\Qb_{kq}(\Ab)$.

Now consider the Hessian of $f(\Ab)$ at a global minimizer $\Ab^*$. The $(k, q)-(k', q')$-th block of the Hessian can be characterized as
\begin{align}
\nabla_{\vecz{\Ab_{k'q'}}} &\left(\nabla_{\vecz{\Ab_{kq}}} f(\Ab^*) \right) \\
&= \nabla_{\vecz{\Ab_{k'q'}}} \left(\Qb_{kq}^\top(\Ab^*) \vecz{\eb} \right) \\
&\overset{(i)}{=}  \left(\nabla_{\vecz{\Ab_{k'q'}}} \Qb_{kq}(\Ab^*) \right)\vecz{\eb}  + \left(\nabla_{\vecz{\Ab_{k'q'}}} \vecz{\eb} \right)\Qb_k(\Ab^*) \\
&\overset{(ii)}{=} 0 + \Qb_{k'q'}(\Ab^*)^\top\Qb_{kq}(\Ab^*),
\end{align}
where (i) follows from the product rule, and (ii) follows from the fact that $\eb(\Wb^*) = 0$, and applies the result to obtain the first order derivative.

\section*{Proof of First and Second Order Derivatives of $g(\Wb)$ in \cref{eq: nonlinear_grad_1,eq: nonlinear_grad_2,eq: nonlinear_hessian}}
For \cref{eq: nonlinear_grad_1}, note that we can rewrite $g(\Wb) = \frac{1}{2} \norm{(\sigma(\Wb_1\Xb)^\top \otimes \Ib)\vecz{\Wb_2} - \vecz{\Yb}}^2$. Then we obtain that 
\begin{align*}
\nabla_{\vecz{\Wb_2}} g(\Wb) = (\sigma(\Wb_1\Xb)^\top \otimes \Ib)^\top \vecz{\eb} = (\sigma(\Wb_1\Xb) \otimes \Ib) \vecz{\eb}.
\end{align*}

For \cref{eq: nonlinear_grad_2}, note that  we can rewrite  $g(\Wb) = \frac{1}{2} \norm{(\Ib \otimes \Wb_2)\vecz{\sigma(\Wb_1\Xb)} - \vecz{\Yb}}^2$. Then, by the composition rule we obtain that 
\begin{align*}
\nabla_{\vecz{\Wb_1}} g(\Wb) &= \left(\nabla_{\vecz{\Wb_1}} \vecz{\sigma(\Wb_1\Xb)} \right) \left(\nabla_{\vecz{\sigma(\Wb_1\Xb)}} g(\Wb) \right) \\
& = \left(\nabla_{\vecz{\Wb_1}}  \sigma((\Xb^\top \otimes \Ib)\vecz{\Wb_1} ) \right) \left(\nabla_{\vecz{\sigma(\Wb_1\Xb)}} g(\Wb) \right),
\end{align*}
where the second term further becomes $\nabla_{\vecz{\sigma(\Wb_1\Xb)}} g(\Wb) = (\Ib\otimes \Wb_2)^\top \vecz{\eb}$. We next derive the following general formula to simplify the first term. 
\begin{align*}
\nabla_\xb \sigma(\Ab\Xb) &= \nabla_\xb [\sigma(\ab_1^\top \xb);\ldots;\sigma(\ab_n^\top \xb)] \\
&=  [\nabla_\xb \sigma(\ab_1^\top \xb);\ldots;\nabla_\xb \sigma(\ab_n^\top \xb)] \\
&= [\sigma'(\ab_1^\top \xb)\ab_1;\ldots; \sigma'(\ab_n^\top \xb)\ab_n ] \\
&= \Ab^\top \mathrm{diag}(\sigma'(\Ab\xb)).
\end{align*}
Thus, applying the above formula with $\Ab = (\Xb^\top \otimes \Ib), \xb = \vecz{\Wb_1}$, the first term becomes
$\nabla_{\vecz{\Wb_1}}  \sigma((\Xb^\top \otimes \Ib)\vecz{\Wb_1} ) = (\Xb^\top \otimes \Ib)^\top \mathrm{diag}(\sigma'(\Xb^\top \otimes \Ib) \vecz{\Wb_1})$. Combining these facts, we further obtain that
\begin{align*}
\nabla_{\vecz{\Wb_1}} g(\Wb) &= (\Xb^\top \otimes \Ib)^\top \mathrm{diag}(\sigma'(\Xb^\top \otimes \Ib) \vecz{\Wb_1}) (\Ib\otimes \Wb_2)^\top \vecz{\eb} \\
&= (\Xb \otimes \Ib) \sigma'(\mathrm{diag}(\vecz{\Wb_1\Xb})) \vecz{\Wb_2^\top\eb} \\
&= (\Xb \otimes \Ib) \vecz{\sigma'(\Wb_1\Xb) \circ (\Wb_2^\top\eb)}.
\end{align*}

To prove \cref{eq: nonlinear_hessian}, we consider the four subblocks of $\nabla_{\vecz{\Wb}}^2 g(\Wb^*)$. For the block $\nabla_{\vecz{\Wb_2}} (\nabla_{\vecz{\Wb_2}} g(\Wb^*))$, we use the formula of $\nabla_{\vecz{\Wb_2}} g(\Wb)$ and obtain that
\begin{align*}
\nabla_{\vecz{\Wb_2}} (\nabla_{\vecz{\Wb_2}} g(\Wb^*)) &= \nabla_{\vecz{\Wb_2}} \big[(\sigma(\Wb_1\Xb) \otimes \Ib) \vecz{\Wb_2 \sigma(\Wb_1\Xb)}\big] \\
&= (\sigma(\Wb_1\Xb) \otimes \Ib)(\sigma(\Wb_1\Xb) \otimes \Ib)^\top.
\end{align*}
For the other diagonal block $\nabla_{\vecz{\Wb_1}} (\nabla_{\vecz{\Wb_1}} g(\Wb^*))$, we use the formula of $\nabla_{\vecz{\Wb_1}} g(\Wb)$ and obtain that 
\begin{align*}
&\nabla_{\vecz{\Wb_1}} (\nabla_{\vecz{\Wb_1}} g(\Wb^*)) \\
&=  \nabla_{\vecz{\Wb_1}} \big[(\Xb \otimes \Ib) \sigma'(\mathrm{diag}(\vecz{\Wb_1 \Xb})) \vecz{\Wb_2^\top \eb}   \big] \\
& \overset{(i)}{=} 0 + \big(\nabla_{\vecz{\Wb_1}} \vecz{\Wb_2^\top \eb}\big) \sigma'(\mathrm{diag}(\vecz{\Wb_1 \Xb}))(\Xb^\top \otimes \Ib) \\
&= \big(\nabla_{\vecz{\Wb_1}} \vecz{\Wb_2^\top \Wb_2 \sigma(\Wb_1 \Xb)}\big) \sigma'(\mathrm{diag}(\vecz{\Wb_1 \Xb}))(\Xb^\top \otimes \Ib) \\
&\overset{(ii)}{=} \big(\nabla_{\vecz{\Wb_1}} \vecz{\sigma(\Wb_1 \Xb)}\big) \big(\nabla_{\vecz{\sigma(\Wb_1 \Xb)}}   \vecz{\Wb_2^\top \Wb_2 \sigma(\Wb_1 \Xb)}\big) \sigma'(\mathrm{diag}(\vecz{\Wb_1 \Xb}))(\Xb^\top \otimes \Ib) \\
&= (\Xb \otimes \Ib) \sigma'(\mathrm{diag}(\vecz{\Wb_1 \Xb})) (\Ib \otimes \Wb_2^\top \Wb_2) \sigma'(\mathrm{diag}(\vecz{\Wb_1 \Xb}))(\Xb^\top \otimes \Ib) \\
&= \big[(\Xb \otimes \Ib) \sigma'(\mathrm{diag}(\vecz{\Wb_1 \Xb})) (\Ib \otimes \Wb_2^\top)\big]\big[(\Xb \otimes \Ib) \sigma'(\mathrm{diag}(\vecz{\Wb_1 \Xb})) (\Ib \otimes \Wb_2^\top)\big]^\top
\end{align*}
where (i) applies the composition rule and uses the fact that $\eb(\Wb^*) = 0$, and (ii) applies the composition rule. 

For the off-diagonal block $\nabla_{\vecz{\Wb_1}} (\nabla_{\vecz{\Wb_2}} g(\Wb^*))$, we obtain that
\begin{align*}
&\nabla_{\vecz{\Wb_1}} (\nabla_{\vecz{\Wb_2}} g(\Wb^*)) = \nabla_{\vecz{\Wb_1}} \big[(\sigma(\Wb_1\Xb) \otimes \Ib) \vecz{\eb}\big] \\
&\overset{(i)}{=} 0 +  \big(\nabla_{\vecz{\Wb_1}} \vecz{\eb}\big) (\sigma(\Wb_1\Xb)^\top \otimes \Ib) \\
&=\big(\nabla_{\vecz{\Wb_1}} \vecz{\Wb_2 \sigma(\Wb_1\Xb)}\big) (\sigma(\Wb_1\Xb)^\top \otimes \Ib) \\
& = \big(\nabla_{\vecz{\Wb_1}} \vecz{\sigma(\Wb_1\Xb)}\big) \big(\nabla_{\vecz{\sigma(\Wb_1\Xb)}}  \vecz{\Wb_2 \sigma(\Wb_1\Xb)}  \big) (\sigma(\Wb_1\Xb)^\top \otimes \Ib) \\
&= (\Xb\otimes \Ib) \sigma'(\mathrm{diag}(\vecz{\Wb_1 \Xb})) (\Ib \otimes \Wb_2^\top) (\sigma(\Wb_1\Xb)^\top \otimes \Ib),
\end{align*}
where (i) applies the composition rule and uses the fact that $\eb(\Wb^*) = 0$. The other off-diagonal block is just the transpose of the above terms. Combining all these expressions of subblocks, it is clear that 
\begin{align}
&\nabla_{\vecz{\Wb}}^2 g(\Wb^*) = \Hb(\Wb^*)^\top \Hb(\Wb^*),~\text{where}~
\Hb^\top = 
\begin{bmatrix}
(\Xb \otimes \Ib)\sigma'(\mathrm{diag}(\vecz{\Wb_1\Xb})) (\Ib\otimes \Wb_2^\top)  \\
\sigma(\Wb_1\Xb) \otimes \Ib 
\end{bmatrix}.
\nonumber
\end{align}

	}

\end{document}